\documentclass[a4paper,twoside,english]{article}

\usepackage{rfia2000}
\usepackage[T1]{fontenc}
\usepackage[utf8]{inputenc}
\usepackage{babel}
\usepackage{times}
\usepackage{amsmath}
\usepackage{amssymb}
\usepackage{graphicx}
\usepackage{ijcai}
\usepackage{comment}

\usepackage{xcolor}
\usepackage{soul}
\usepackage{footnote}
\usepackage[para]{threeparttable}

\makesavenoteenv{tabular}
\makesavenoteenv{table}

\newcommand{\argmax}[1]{\underset{#1}{\operatorname{arg}\,\operatorname{max}}\;}
\definecolor{aqua}{rgb}{0.0, 1.0, 1.0}

\newcommand{\citename}[1]{\citeauthor{#1} (\citeyear{#1})}

\begin{document}
%%%%%Pas de date
\date{}
%%%%% Titre gras 14 points
\title{\Large\bf A survey on intrinsic motivation in reinforcement learning}
%%%%% Si auteur unique
%\author{L. Auteur \\
%%  Son institut \\
%%  Son addresse \\
%%  Son email}

%%%% pour deux auteurs
%\author{\begin{tabular}[t]{c@{\extracolsep{8em}}c}
%%%% pour trois auteurs
\author{\begin{tabular}[t]{c@{\extracolsep{6em}}c@{\extracolsep{6em}}c}
%%%% pour quatre auteurs
%%%%\author{\begin{tabular}[t]{c@{\extracolsep{4em}}c@{\extracolsep{4em}}c@{\extracolsep{4em}}c}
%%%%pour plus d\'ebrouillez-vous !
A. Aubret${}^1$  & L. Matignon${}^1$  & S. Hassas${}^1$\\
\end{tabular}
{} \\
 \\
${}^1$        Univ Lyon, Université Lyon 1, CNRS, LIRIS, F-69622, Villeurbanne, France  
{} \\
 \\
%Mon adresse compl\`ete \\
arthur.aubret@liris.cnrs.fr\\
%{\bf Domaine principal de recherche}: RFP ou IA\\
%{\bf Papier soumis dans le cadre de la journée commune}: OUI ou NON
}

\maketitle

\begin{abstract}

The reinforcement learning (RL) research area is very active, with an important number of new contributions; especially considering the emergent field of deep RL (DRL). However a number of scientific and technical challenges still need to be addressed, amongst which we can mention the \textit{ability to abstract actions} or \textit{the difficulty to explore the environment} which can be addressed by intrinsic motivation (IM). In this article, we provide a survey on the role of intrinsic motivation in DRL. We categorize the different kinds of intrinsic motivations and detail for each category, its advantages and limitations with respect to the mentioned challenges. Additionnally, we  conduct an in-depth investigation of substantial current research questions, that are currently under study or not addressed at all in the considered research area of DRL. We choose to survey these research works, from the perspective of \textit{learning how to achieve tasks}. We suggest then, that solving current challenges could lead to a larger developmental architecture which may tackle most of the tasks. We describe this developmental architecture on the basis of several building blocks composed of a RL algorithm and an IM module compressing information.

\end{abstract}

\section{Introduction}

In reinforcement learning (RL), an agent learns by trial-and-error to maximize the expected rewards gathered as a result of its actions performed in the environment \cite{sutton1998reinforcement}. Traditionally, an agent maximizes a reward defined according to the task to perform: it may be a score when the agent learns to solve a game or a distance function when the agent learns to reach a goal. The reward is then considered as extrinsic (or as a feedback) because the reward function is provided expertly and specifically for the task. With an extrinsic reward, many spectacular results have been obtained on Atari game \cite{bellemare15} with the Deep Q-network (DQN) \cite{mnih2015human} through the integration of deep learning to RL, leading to deep reinforcement learning (DRL). However, these approaches turn out to be most of the time unsuccessful when the rewards are %too sparse
scattered in the environment, as the agent is then unable to learn the desired behavior for the targeted task \cite{franccois2018introduction}. 

Moreover, the behaviors learned by the agent are hardly reusable, both within the same task and across many different tasks \cite{franccois2018introduction}. It is difficult for an agent to generalize its skills so as to learn to make high-level (or abstract) decisions in the environment. For example, such abstract decision could be \textit{go to the door} using primitive actions (or low-level actions) consisting in moving in the four cardinal directions; or even to \textit{move forward} controlling different joints of a humanoid robot like in the robotic simulator MuJoCo \cite{todorov2012mujoco}. %Such abstract decisions are often called options \cite{sutton1999between}.
Such abstract decisions, often called options \cite{sutton1999between}, have to be learned, but there is potentially an infinite number of options in real-world-like simulator and some are more complex than others. It follows that an agent must learn options in a specific order. For example, a robot should learn to grasp an object before learning to put it into a box; it should also learn to reach the kitchen door before learning to reach the sink from the bedroom. In fact this is an exploration problem in the space of options rather than in the space of states (as described above). Therefore if the agent does not consider the order of tasks, its learning process will take longer than if he took the order into consideration. This issue is currently studied by\textit{curriculum learning}\cite{bengio2009curriculum}. 

%In addition, it appears that classical DRL algorithms as well as the above-mentioned unresolved issues could strongly benefit from a good state representation \cite{raffin2019decoupling} (see \S \ref{motiv}). \\
A common limitation of state-of-the-art DRL algorithms, and their extensions tackling the above-mentioned issues, is their inability to learn a shared state representation independently from the extrinsic rewards. Nevertheless, learning on interesting features rather than the ground state space allows DRL algorithms to considerably speed up the learning process \cite{raffin2019decoupling}. For example, this is easier to learn to navigate in an environment using coordinates rather than with first-person images.

On another side, unlike RL, developmental learning \cite{piaget1952origins,cangelosi2018babies,oudeyer2016evolution} is based on the trend that babies, or more broadly organisms, have to spontaneously explore their environment \cite{gopnik1999scientist,georgeon2011early} and acquire new skills \cite{barto2013intrinsic}. This is commonly called intrinsic motivation (IM), which can be derived from an intrinsic reward. This kind of motivation allows to gain autonomously new knowledge and skills, which then makes the learning process of new tasks easier \cite{baldassarre2013intrinsically}.

For several years now, IM is increasingly used in RL, fostered by important results and the emergence of deep learning. This paradigm offers a greater learning flexibility, through the use of a more general reward function, allowing to tackle the issues raised above when only an extrinsic reward is used. Typically, IM improves the agent ability to explore its environment, to incrementally learn skills (options) independently of its main task, to choose an adequate skill to be improved and even to create a representation of its state with meaningful properties. In addition, as a consequence of its definition, IM does not require additional expert supervision, making it easily generalizable across environments. \\

In this article, we investigate the use of IM in the framework of DRL and consider the following aspects:
\begin{itemize}
    \item Characteristics of IM.
    \item IM in the framework of DRL.
    \item Role of IM in addressing the above mentioned challenges.
    \item Actual limitations of the use of IM in RL, and the associated challenges.
\end{itemize}
%An other contribution of this article is to provide an unified view of the state-of-the-art based on  compression theory. Moreover, we propose an in-depth analysis of intrinsic motivation in DRL linking these methods to developmental learning. Specifically, we propose a general developmental architecture unifying all the approaches and highlighting the numerous perspectives in this domain.
An other contribution of this article is to provide an unified view of the state-of-the-art based on compression theory. Moreover, we propose an analysis of approaches using IM in DRL and their relation to developmental learning. More specifically, we propose a general developmental architecture unifying all the approaches and highlighting the numerous perspectives in this domain.

Our study is not meant to be exhaustive. It is rather a review of current ongoing research directions, their limitations and their potential perspectives. The overall literature on IM is huge \cite{barto2013intrinsic} and our review only considers its application to DRL. We highlight how IM can improve over state of the art DRL algorithms, scaling to large state and action dimension spaces. 

A recent study on IM has recently been achieved by \citename{linke2019adapting} that is complementary to ours. This study focus on IM in the context of active learning but only study IM for \textit{curriculum learning}. In our paper, we focus on works related to DRL, where we investigate a larger spectrum of IM kinds. We also provide a thorough analysis of works, and identify outlooks of the domain.\\

This survey paper is organized as follows. As a first step, we introduce the basic concepts used in the rest of the paper,  
%define the key elements of the article, 
namely Markov decision processes, goal-parameterized RL, the bases of information theory, intrinsic motivation and \textit{empowerment} (Section \ref{defs}). In Section \ref{sect:defis}, we highlight the main current challenges of RL and identify the need for an additional outcome. This brings us to explain how to combine IM and RL and how to classify different sorts of IM (Section \ref{sec:embedded_rl}). Then we detail the work integrating RL and IM by first studying articles relying on knowledge acquisition (Section \ref{motiv}) and secondly those based on skills learning (Section \ref{skill_learning}). Thereafter, we emphasize the current challenges of these models (Section \ref{limite}) and identify achievements and issues of classes of algorithm in regard to exigent tasks (Section \ref{tasks}). Finally, we take a step back and analyze common aspects to those methods and propose their integration in a developmental learning framework (Section \ref{analyse}).

\section{Definitions and Background}\label{defs}

In this section, we will review the background of RL field and its recent extension through goal-parameterized RL. We will then introduce some fundamentals of information theory and explain the concept of IM. We will then introduce the theoretical definition of \textit{empowerment}, used as IM.%be able to give the theoretical definition of an important intrinsic motivation which is the \textit{empowerment}.

\subsection{Markov decision process}
%Cadre formel du RL et définition du MDP
The goal of a Markov decision process (MDP) is to maximize the expectation of cumulative rewards received through a sequence of interactions. It is defined by:
\begin{itemize}
\item $S$ the set of possible states;
\item $A$ the set of possible actions;
\item $P$ the transition function $P : S \times A \times S \rightarrow \mathbb{P}(S'|S,A)$:
\item $R$ the reward function $R : S \times S \times A \rightarrow \mathbb{R}$;
\item $\gamma \in[0,1]$ the discount factor;
\item $\rho_0 : S \rightarrow \mathbb{P}(S)$ the initial distribution of states.
\end{itemize}

An agent starts in a state $s_0$ given by $\rho_0$. At each time step $t$, the agent is in a state $s_t$ and performs an action $a_t$; then it waits for the feedback from the environment composed of a state $s_{t+1}$ sampled from the transition function $P$, and a reward $r_t$ given by the reward function $R$. The agent repeats this interaction loop until the end of an episode. The goal of an MDP is to maximize the long-term reward defined by:
\begin{equation}
\left[\sum_{t=0}^{\infty} \gamma^t r_t\right] .
\end{equation}

A reinforcement learning algorithm aims to associate actions $a$ to states $s$ through a policy $\pi$. The goal of the agent is then to find the optimal policy $\pi^*$ maximizing the reward:
\begin{equation}
\pi^* = \argmax{\pi} \mathbb{E} \left[\sum_{t=0}^{\infty} \gamma{^t} R(s{_t},s_{t+1},\pi(s{_t}))\right] .
\end{equation}

In order to find the action maximizing the long-term reward in a state $s$, it is common to maximize the expected discounted gain following a policy $\pi$ from a state, noted $V_{\pi}(s)$, or from a state-action tuple, noted $Q_{\pi}(s,a)$ (cf. Equation \eqref{eq:espeQ}). It enables to %tackle the \textit{credit assignment problem} by measuring 
measure the impact of the state-action tuple in obtaining the %cumulated 
cumulative reward \cite{sutton1998reinforcement}. 
\begin{equation}
Q_{\pi}(s,a) = E_{a{_t}\sim\pi(s{_t})} \left(\sum_{t=0}^{\infty} \gamma{^t} R(s{_t},a{_t})|_{s_0=s,a_0=a} \right).
\label{eq:espeQ}
\end{equation}

To compute these values, it is possible to use the Bellman equation \cite{sutton1998reinforcement}:
\begin{equation}
\label{eq:bellman}
Q_{\pi}(s_t,a_t) = R(s_t,a_t) + \gamma Q_{\pi}(P(s_t,a_t),a_{t+1}).
\end{equation}

$Q$ and/or $\pi$ are often approximated with neural networks when the state space is continuous or very large \cite{mnih2016asynchronous,lillicrap2015continuous}.

\subsection{Goal-parameterized RL}\label{uvfa}

Usually, RL is used to solve only one task and is not suited to learn multiple tasks. Typically, an agent is unable to generalize across different variants of a task. For instance, if an agent learns to grasp a circular object, it will not be able to grasp a square object. One way to generalize DRL to multi-goal learning, or even to every available goal in the state space, is to use the universal value function approximator (UVFA) \cite{schaul2015universal}. It should be noted that each state can serve as a target goal. Let us consider an agent moving in a closed maze where every position in the maze can be a goal. Assuming that there exists a vector space where a goal has a representation, UVFA integrates, by concatenating, the state goal representation with the observation of the agent. The found policy is then conditioned on the goal: $\pi(s)$ becomes $\pi(s,g)$ where $g$ is a goal. It involves that if the goal space is well-constructed (as a state space for example), the agent can generalize its policy across the goal space. A similar idea can be retrieved with contextual policy search \cite{fabisch2014active}.

When the goal space is exactly a continuous state space, it is difficult to determine whether a goal is reached or not, since two continuous values are never exactly equal. Hindsight experience replay (HER) \cite{andrychowicz2017hindsight} tackles this issue by providing a way to learn on multiple objectives with only one interaction. With author's method, the agent can use an interaction done to accomplish one goal to learn on an other goal, by modifying the associated reward. 
Let us roll out an example, an agent acts in the environment, resulting in an interaction $(s,s',r_g,a,g)$ where $r_g$ is the reward associated to the goal $g$. The agent can learn on this interaction, but can also use this interaction to learn other goals; to do so, it can change the goal into a new goal and recompute the reward, resulting in a new interaction $(s,s',r_{g'},a,g')$. The only constraint for doing this is that the reward function $R(s,a,s',g')$ has to be known. Typically, an agent can have a goal state and a reward function which is $1$ if it is into that state and $0$ otherwise. At every interaction, it can change its true goal state for its current state and learn with a positive reward.              

%Indeed, using only one interaction with the environment $(s,s',r_g,a,g)$ with $r_g$ the reward associated to the goal $g$, it is possible to create a new interaction with a new objective (which would ideally be successful) $(s,s',r_{g'},a,g')$ as long as the reward function $R(s,a,s',g')$ is available.

%Several strategies can be used to sample the new goals $g'$ %\cite{bai2019guided,ren2019exploration,andrychowicz2017hindsight}. Such complex strategies can improve the policy diversity and exploration through a \textit{curriculum}; even though some standard sampling strategies are out of our scope, in \S\ref{sec:curriculum}, we will focus our study on strategies based on reinforcement learning.%since we only study intrinsic motivation as a reinforcement signal.

\subsection{Information theory}
%Brève introduction à l'information de shannon, l'entropie et l'information mutuelle, 

The Shannon entropy quantifies the mean necessary information to determine the value of a random variable. Let $X$ be a random variable with a law of density $p(X)$ satisfying the normalization and positivity requirements, we define its entropy by:
\begin{equation}
H(X) = -\int_{X} p(x)\log p(x) .
\end{equation}

In other words, it allows to quantify the disorder of a random variable. The entropy is maximal when $X$ follows an uniform distribution, and minimal when $p(X)$ is equal to zero everywhere except in one value, which is a Dirac distribution. From this, we can also define the entropy conditioned on a random variable $S$. It is similar to the classical entropy and quantifies the mean necessary information to find $X$  knowing the value of an other random variable $S$:
\begin{equation}
H(X|S) = -\int_{S} p(s)\int_{X} p(x|s)\log p(x|s).
\end{equation}

The mutual information allows to quantify the information contained in a random variable $X$ about an other random variable $Y$. It can also be viewed as the decrease of disorder brought by a random variable $Y$ on a random variable $X$. The mutual information is defined by:
\begin{equation}
I(X;Y) =  H(X) - H(X|Y)
\end{equation}

We can notice that the mutual information between two independent variables is zero (since $H(X|Y)=H(X)$). Similarly to the conditional entropy, the conditional mutual information allows to quantify the information contained in a random variable about an other random variable, knowing the value of a third one. It can be written in various ways:
\begin{align}
    I(X;Y|S) &= H(X|S) - H(X|Y,S)\label{information}\\ 
    &= H(Y|S) - H(Y|X,S)   \label{information2} \\
    &= H(X|S) + H(Y|S) - H(X,Y|S) \nonumber \\
    &= D_{KL} \Big[ p(x,y|s) || p(x|s)p(y|s)\Big] \label{kldiv} 
\end{align}

We can see with equations \eqref{information} and \eqref{information2} that the mutual information is symmetric and that it characterizes the decrease in entropy on X brought by Y (or inversely). Equation \eqref{kldiv} defines the conditional mutual information as the difference between distribution $P(Y,X|S)$ and the same distribution if $Y$ and $X$ were independent variables (the case where $H(Y|X,S) = H(Y|S)$). For further information on these notions, the interested reader could refer to \cite{tishby2000information,ito2016information,cover2012elements}.

\subsection{Intrinsic motivation}

The idea of IM is to push an agent to get a specific behavior without any direct feedback from the environment. Simply stated, it is about doing something for its inherent satisfaction rather than to get a reward assigned by the environment \cite{ryan2000intrinsic}. This kind of motivation comes from developmental learning, which is inspired by the trend of babies to develop skills by exploring their environment \cite{gopnik1999scientist,white1959motivation}.

More rigorously, \citename{oudeyer2008can} explain that an activity \textit{is intrinsically motivating for an autonomous entity if its interest depends primarily on the collation or comparison of information from different stimuli and independently of their semantics}. The main point is that the agent must not have any \textit{a priori} on the semantic of the observations it receives. We notice that the term of comparison of information refers directly to information theory defined previously. At the opposite, an extrinsic reward results of an unknown environment static function which does not depend on previous experience of the agent on the considered environment.

Typically, a student doing his mathematical homework because he thinks it is interesting is intrinsically motivated whereas his classmate doing it to get a good grade is extrinsically motivated. The concept of \textbf{intrinsic/extrinsic} refers to the \textit{why of the action}, this should not be confused with internality/externality which refers to the location of the reward \cite{oudeyer2008can}.

Table \ref{tab:rlim} shows the difference between reinforcement learning and the use of IM. Reinforcement learning is an active process since the agent learns from its interactions with the environment, unlike classification or regression which are supervised methods. Unsupervised learning is a passive learning process, \textit{i.e.} it does not use predefined labels, or in other words, learns without a feedback. Finally, the substitution of the feedback by an intrinsic reward allows to break free from an expert supervision; however, the difference remains between IM and unsupervised learning in the sense that IM is an active process which implies interactions.

An extensive overview of IM, beyond the RL framework, can be found in \citename{barto2013intrinsic}.
\begin{table}[t]
\centering

 \caption{Type of learning. \textit{feedback} here refers to an expert supervision.}\label{tab:rlim}
\begin{tabular}{|l|l|l|}
  \hline
   & With \textit{feedback}  & Without \textit{feedback} \\
  \hline
    Active & Reinforcement & Intrinsic motivation \\
    Passive & Supervised & Unsupervised \\
  \hline

\end{tabular}
\end{table}

\subsection{\textit{Empowerment}}\label{sect:background_empowerment}

The notion of \textit{empowerment} has been developed to answer the following question: is there a local utility function which makes possible the survival of an organism \cite{klyubin2005empowerment,salge2014empowerment}? This hypothetical function should be local in the sense that it does not modify the organism behavior on the very long term (death itself does not impact this function) and induced behaviors have to help species survival. Typically, this function can explain animal's will to dominate its pack, and more generally, the human's wish to acquire a social status, to earn more money or to be stronger, the need to maintain a high blood sugar level or the fear to be hurt \cite{klyubin2005empowerment,salge2014changing}. Each of these motivations widens the possibilities of actions of the agent, and thereby its influence: a person with many resources will be able to do more things than a poor one. \citename{klyubin2005empowerment} named this ability to control the environment the \textit{empowerment} of an agent.

The \textit{empowerment} is usually defined using information theory. \citename{klyubin2005empowerment} interpret the interaction loop as the sending of information into the environment: an action is a signal being sent while the observation is a received signal. %The more informative the action is about the next observations, the more the \textit{empowerment} will be high. 
The more informative the action about the next observations, the more the \textit{empowerment}.
{Empowerment} is measured as the capacity of a channel linking the actions and observations of the agent. Let $a_t^n = (a_t,a_{t+1},...,a_{t+n})$ be the actions executed by the agent from time $t$ to $t+n$, and $s_{t+n}$ the state of the environment at the time step $t+n$. The \textit{empowerment} of state $s_t$, noted $\Sigma(s_t)$, is then defined as:
\begin{align}
    \Sigma(s_t) &= \max_{p(a_t^n)}I(a_t^n;s_{t+n}|s_t) \nonumber \\
    &= \max_{p(a_t^n)}H(a_t^n|s_t)-H(a_t^n|s_{t+n},s_t). \label{eq:meaning}
\end{align}

Maximizing the \textit{empowerment} is the same as looking for the state in which the agent has the most control on the environment. Typically, the second term of Equation \ref{eq:meaning} allows the agent to be sure of where he is going, whereas the first term emphasizes the diversity of reachable states. To get a large overview on the different ways to compute the \textit{empowerment}, the reader can refer to \citename{salge2014empowerment}. Hereafter in this article, we will focus on the application of \textit{empowerment} in the context of RL, that is why we will not detail work using \textit{empowerment} out of the RL context (see e.g. \citename{karl2017unsupervised}, \citename{guckelsberger2016intrinsically}, \citename{capdepuy2007maximization}, \citename{salge2014empowerment}).

%\section{Intrinsic motivation embedded into RL}\label{MI+RL}
\section{Challenges of RL tackled with IM}\label{sect:defis}

%In this section, we detail the main challenges of reinforcement learning that can be addressed with intrinsic motivation. 
In this section, we identify four challenges in DRL for which IM provides a suitable solution. We illustrate these challenges and explain their importance.

%\subsection{RL problematic}\label{sect:defis}
\subsection{Sparse rewards} 

\begin{figure*}
\begin{centering}
\includegraphics[width=10cm]{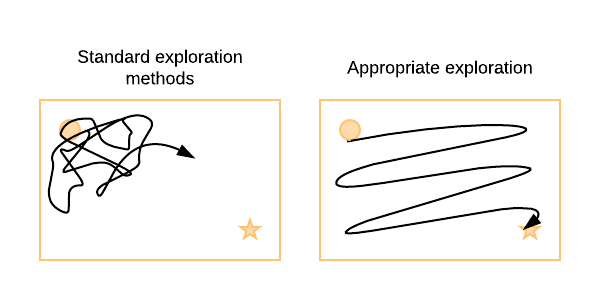}
\caption{Illustration of the sparse reward issue in a very simple setting. The agent, represented by a circle, strives to reach the star. The reward function is one when the agent reaches the star and zero otherwise. On the left side, the agent explores with standard methods such as $\epsilon-greedy$; as a result, it stays in its surrounded area because of the temporal inconsistency of its behaviour. On the right side, we can imagine an ideal exploration strategie where the agent covers the whole state space to discover where rewards are located.}
\label{im:sparse_reward}
\end{centering}
\end{figure*}

Classic RL algorithms operate in environments where the rewards are \textbf{dense}, \textit{i.e.} the agent receives a reward after almost every completed action. In this kind of environment, naive exploration policies such as $\epsilon$-greedy \cite{sutton1998reinforcement} or the addition of a Gaussian noise on the action \cite{lillicrap2015continuous} are effective. More elaborated methods can also be used to promote exploration, such as Boltzmann exploration \cite{cesa2017boltzmann,mnih2015human}, an exploration in the parameter-space \cite{plappert2017parameter,ruckstiess2010exploring,fortunato2017noisy} or Bayesian RL \cite{ghavamzadeh2015bayesian}. In environments with \textbf{sparse} rewards, the agent receives a reward signal only after it executed a large sequence of specific actions. The game \textit{Montezuma's revenge} \cite{bellemare15} is a benchmark illustrating a typical sparse reward function. In this game, an agent has to move between different rooms while picking up objects (it can be keys to open doors, torches, ...). The agent receives a reward only when it finds objects or when it reaches the exit of the room. Such environments with sparse rewards are almost impossible to solve with the above mentioned exploration policies since the agent does not have local indications on the way to improve its policy. Thus the agent never finds rewards and cannot learn a good policy with respect to the task \cite{mnih2015human}. Figure \ref{im:sparse_reward} illustrates the issue on a simple environment.% where the agent strives to find a star. 

Rather than working on an exploration policy, it is common to shape an intermediary dense reward function which adds to the reward associated to the task in order to make the learning process easier for the agent \cite{su2015reward}. However, the building of a reward function often reveals several unexpected errors \cite{ng1999policy,amodei2016concrete} and most of the time requires expert knowledge. For example, it may be difficult to shape a local reward for navigation tasks. Indeed, one has to be able to compute the shortest path between the agent and its goal, which is the same as solving the navigation problem. On the other side, the automation of the shaping of the local reward (without calling on an expert) requires too high computational resources \cite{chiang2019learning}.

We will see in \S\ref{curiosity} how IM is a valuable method to encourage exploration in a sparse rewards setting. In \S\ref{sec:curriculum}, we also provide details on the value of IM in the context of\textit{curriculum learning}for exploration.

%--P3
\subsection{Building a good state representation}\label{sec:staterepresentation}

What is a good state representation? \citename{bohmer2015autonomous} argue that, in standard RL, this representation must be markovian, able to represent the true value of the policy, generalize well and low-dimensional. Using such adapted feature space to learn a task can considerably accelerate the learning process \cite{raffin2019decoupling,de2018integrating} and may even help with other computations such as learning a forward model. The best way to do this may be to construct a minimal feature space with \textbf{disentangle features} \cite{bengio2013representation,lesort2018state}. 

In order to better understand the importance of a relevant state representation in RL, let us consider a simple navigation task where the agent has to reach a target area in an empty space. If the agent accesses pixels input from above, it will have to extract its own position and the target position through complex non-linear transformations to understand which directions it has to take. At the opposite, if it has already access to its position, it will only have to check if its vertical and horizontal positions are greater, equals or smaller than those of the target. In standard RL, this problem is exacerbated, firstly because the only available learning process is the back-propagation of the reward signal, and secondly by the presence of noise in the raw state. It results that if the reward is sparse, the agent will not learn anything from its interactions even though interaction by themselves are rich in information. Furthermore, the state representation learned with a reward fully depends on the task and cannot be generalized to other tasks, whereas a state representation learned independently from the task can be used for other tasks.

Several works are about the learning of a relevant state representation. Auxiliary losses can complement the reward with supervised learning losses. It relies on information such as immediate reward or other predefined functions \cite{shelhamer2016loss,jaderberg2016reinforcement}. The agent may also use some prior knowledge on transitions \cite{jonschkowski2015learning,jonschkowski2017pves} or learn inverse models \cite{zhang2018decoupling}. There is a large literature on the best way to quickly build this kind of state space, we invite the interested reader to look at \cite{lesort2018state} for a general review and recommend \cite{bengio2013representation} for an introduction to the learning of representations. However, it is still difficult to get an entire disentangled representation of controllable objects since it can require interactions with the environment.

Although this issue did not attracted much attention, we will exhibit in Section \ref{sec:staterep} how IM can be a key component in order to build a state representation with such meaningful properties. We emphasize that we focus on works for which the intrinsic goal of the agent is to learn a \textit{state representation}. As a consequence, other ways to learn a \textit{state representation} are out of the scope of the section.

\subsection{Temporal abstraction of actions} \label{sec:abstraction}

Temporal abstraction of actions consists in using high-level actions, also called \textbf{options}, which can have different execution times \cite{sutton1999between}. Each option is associated with an \textbf{intra-option policy }which defines the action (low-level actions or other options) to realize in each state when the option is executed. The length of an option, which is the number of executed actions when an option is chosen, is often fixed. An \textbf{inter-option policy} can be in charge of choosing the options to accomplish. Abstract actions are a key element to accelerate the learning process since the number of decisions to take is significantly reduced if options are used. It also makes easier the \textit{credit assignment problem} \cite{sutton1998reinforcement}.
This problem refers to the fact that rewards can occur with a temporal delay and will only very weakly affect all temporally distant states that have preceded it, although these states may be important to obtain that reward. Indeed, the agent must propagate the reward along the entire sequence of actions (through Equation \eqref{eq:bellman}) to reinforce the first involved state-action tuple. This process can be very slow when the action sequence is large. This problem also concerns determining which action is decisive for getting the reward. %Figure \ref{im:abstract_actions} illustrates the issue. Let us assume that agent circles are trying to reach a star which is far from them. If the agent (our green agent) has an option \texttt{Go to the far right} and follows it, he will be rewarded at the end of its option. Then it will be easy to associate the success of its move to its option \texttt{Go to the far right}. In contrast, if the robot has to learn every cardinal move he has to do (low-level or primitives actions), it will be harder to determine which action is responsible for reaching the star, among all executed actions. Furthermore, using options can make exploration easier when rewards are sparse. In figure \ref{im:abstract_actions}, the problem of exploration becomes trivial for the green agent, since only one exploration action can lead to the reward. For the orange agent, it requires an entire sequence of specific low-level actions. This problem arises from the minimal number of actions needed to get a reward.

For example, let us assume that a robot is trying to reach a cake on a table which is far from the robot. If the robot has an option \texttt{get to the table} and follows it, the robot will then only have to take the cake to be rewarded. Then it will be easy to associate the acquisition of the cake (the reward) to the option \texttt{get to the table}. In contrast, if the robot has to learn to handle each of its joints (low-level or primitives actions), it will be difficult to determine which action is responsible of the acquisition of the cake, among all executed actions.%every actions executed. 

\begin{figure*}
\begin{centering}
\includegraphics[width=7cm]{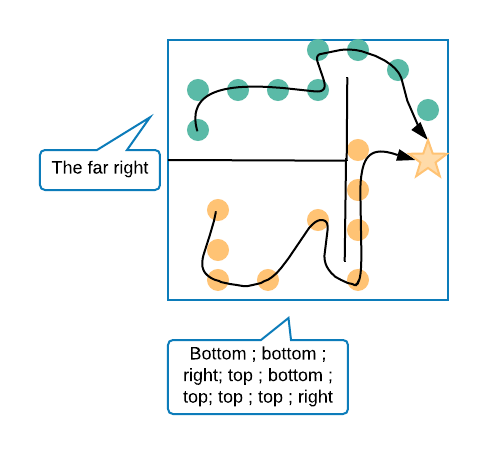}
\caption{Illustration of the benefits of using \textit{options}. Agents, represented by circles, have to reach the star. The green agent can use an \textit{option} \texttt{Go to the far right}; the orange agent can only use primitive actions to reach the star.}
\label{im:abstract_actions}
\end{centering}
\end{figure*}

Furthermore, using options can make exploration easier when rewards are sparse, as illustrated in Figure \ref{im:abstract_actions}. %The green circle can use an option \texttt{Go to the far right}, to reach the rewarding star while the orange agent can only use cardinal movements. 
The problem of exploration becomes trivial for the agent using options, since one exploration action can lead to the reward, yet it requires an entire sequence of specific low-level actions for the other agent. This problem arises from the minimal number of actions needed to get a reward. A thorough analysis of this aspect can be found in \cite{nachum2019does}.%Furthermore, using options can make exploration easier when rewards are sparse. To illustrate this, let's assume that the agent has access to the option \texttt{get the key} in \textit{Montezuma's revenge}. The problem becomes trivial since only one exploration action can lead to the reward, yet it would require without options an entire sequence of specific low-level actions. This problem arises from the minimal number of actions needed to get a reward.

Regarding the intra-option policy, it can be manually defined, but it requires some extra expert knowledge \cite{sutton1999between}. It can also be learnt with the reward function \cite{bacon2017option,riemer2018learning}, but then, options are not reusable for other tasks and are helpless for the exploration problem.

In Section \ref{gen_goal}, we investigate how IM can bring new insights in handling this.

\subsection{Building a curriculum}

Curriculum learning commonly takes place in the framework of multi-task reinforcement learning \cite{WilsonFRT07,LiLC09} where one agent tries to solve several tasks. This is about defining a schedule in the learning process. It comes from the observation that learning is much easier when examples or tasks are organized in a meaningful order \cite{bengio2009curriculum}. Typically, a curriculum could organize tasks in such a way that they are increasingly complex and close to each other. For example, an helpful curriculum may be to first learn to a robot how to grasp a cube and only then how to move the cube; this way, the robot can take advantage of its ability to grasp a cube to move it. Without any prior knowledge, a robot would probably never succeed in grasping and moving a cube since it requires a large sequence of actions (if the robot moves its joints).

Standard methods rely on pre-specified tasks sequences as a curriculum \cite{karpathy2012curriculum}, or expert score which acts as a baseline score \cite{SharmaR17}. Some other methods require strong assumptions \cite{FlorensaHWZA17}, rely on task decomposition \cite{WuZS18} or availability of source tasks \cite{SvetlikLSSWS17,riedmiller2018learning}. It follows that most of the time in standard methods,\textit{curriculum learning}requires an expert in one way or another.

At the opposite, we will demonstrate in Section \ref{sec:curriculum} that it is possible to replace expert knowledge with IM to both speed up multi-task learning and indirectly make exploration easier.

\subsection{Summary}

In summary, several issues in RL are entirely or partially unsolved:
\begin{description}
\item[Sparse rewards:] The agent never reaches a reward signal in case of sparse rewards.
\item[State representation:] The agent does not manage to learn a representation of its observations with independent features or meaningful distance metrics.
\item[Building option:] The agent is unable to learn abstract high-level decisions independently from the task. 
\item[Learning a curriculum:] The agent hardly defines a curriculum among its available goals without expert knowledge. 
\end{description}
All these problems have a common source: reinforcement learning originally tries to solve everything with the extrinsic reward which is a poor source of information. Thereby, it seems relevant to take advantage of other outcomes. We will see in the following (Sections \ref{motiv} and \ref{skill_learning}) how these issues are currently tackled by IM.

%\subsection{A new model of RL with intrinsic rewards}\label{sec:modelRL}
\section{Intrinsic motivation embedded into RL}\label{sec:embedded_rl}

In this section, we describe how IM can be theoretically integrated to a RL framework. Then, we then explain how we categorized works using IM in RL.

\subsection{A new model of RL with intrinsic rewards}\label{sec:modelRL}

\begin{figure*}
\begin{centering}
\includegraphics[width=8.5cm]{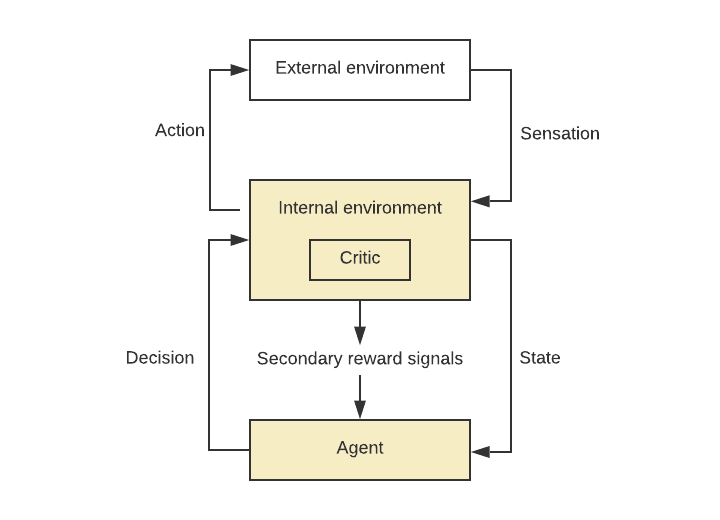}
\caption{New model of RL integrating IM, adapted from \protect\citename{singh2010intrinsically}. }
\label{im:rlintrinsic}
\end{centering}
\end{figure*}

Reinforcement learning is derived from behaviorism \cite{skinner} and uses extrinsic rewards \cite{sutton1998reinforcement}. However \citename{singh2010intrinsically} and \citename{barto2004intrinsically} reformulated the RL framework to incorporate IM. These authors distinguish primary reward signals and secondary reward signals. The secondary reward signal is a local reward computed through expected future rewards and is related to the value function (cf. eq. \eqref{eq:espeQ}) whereas the primary reward signal is the standard reward signal. They differentiate rewards, which are events in the environment, and reward signal which are internal stimulus to the agent. Rather than considering the MDP environment as the environment in which the agent achieves its task, they suggest that the MDP environment can be formed of two parts: the \textbf{external part} which corresponds to the task environment of the agent; the \textbf{internal part} which is internal to the agent and computes the MDP states and the secondary reward signal though previous interactions. Consequently, we can consider an intrinsic reward as a reward received from the MDP environment. The MDP state is no more the external state but an internal state of the agent; it then contradicts what was previously thought as being a limitation of RL \cite{georgeon2015modeling}. In the following, we will use the term of \textit{reward} to name a reward signal. Figure \ref{im:rlintrinsic} summarizes the new framework: the critic is the internal part which computes the intrinsic reward and deals with the credit assignment. The state includes sensations and potentially the history of agent's interactions. The decision can be a high-level decision decomposed into low-level actions.

According to \citename{singh2010intrinsically}, evolution provides a general intrinsic reward function which maximizes a fitness function. We think that such IM can be a meta-skill facilitating the learning of other behaviors. Curiosity, for instance, does not immediately produce selective advantages but enables the acquisition of skills providing by themselves some selective advantages. More widely, the use of IM enables to obtain intelligent behaviors which can serve goals more efficiently than with only a standard reinforcement \cite{lehman2008exploiting} (see Section \ref{motiv}).

In practice, there are multiple ways to integrate an intrinsic reward into a RL framework. The main approach is to compute the agent's reward $r$ as a weighted sum of an intrinsic reward $r_{int}$ and the extrinsic reward $r_{ext}$: $r=\alpha r_{int} + \beta r_{ext}$ \cite{burda2018exploration,gregor2016variational,vezhnevets2017feudal,huang2019learning}. In this version, one can think of the intrinsic reward as an intrinsic bonus. When the extrinsic value function is important to compute the intrinsic reward or when the hyper-parameters have to be different, the sum can be made on the value function level, i.e. $V(s)=\alpha V_{int}(s)+\beta V_{ext}(s)$ \cite{kim2019curiosity}.
Let us notice that when intrinsic rewards evolve over time, the agent generally cannot find an optimal stationary policy. 

A less common way to integrate it can be with sampling methods. When a goal space is available to the agent, it can choose what to do independently from the task. This method is more specific to works on curriculum learning, thereby we will provide more details in Section \ref{sec:curriculum}.
%Another possibility is to consider the option framework. For example, \citename{kulkarni2016hierarchical} or \citename{eysenbach2018diversity} make use of intrinsic reward to learn some skills, and use extrinsic reward (or an other intrinsic reward as in \citename{fournier2019clic}) to choose the skill to use. The main advantage in decoupling options from the extrinsic reward is to have options independent of the task, thus the options can be reused for an other task (see Section \ref{gen_goal}). 

%Instead of learning a task or an option, it is also possible to use an intrinsic reward to learn a transition model \cite{2019arXiv190509275W} or a state representation \cite{thomas2018disentangling,ghosh2018learning}.

\subsection{Classification of the use of IM in RL}

\citename{oudeyer2008can} already proposed a classification of the different IMs where the two major models are either knowledge-based or competence-based. The first one consists of a comparison between agent's predictions and reality, and the second one refers to the performance on self-generated goals. %We found that to be inadequate relatively to applications in RL and propose a slight modification. 
We propose a slightly different classification to include skill abstraction and highlight skill acquisition. Our classification emphasizes two major kinds of IM in RL and is summarized in the Table \ref{tab:tableofcontents}.
\begin{description}
\item[Knowledge acquisition]: With this motivation, the agent strives to find new knowledge about its environment. This knowledge can concern what it can/cannot control, the functioning of the world, discovering new areas or understanding the sense of proximity. It is very close to the knowledge-based classification of \citename{oudeyer2008can}. We will see that: 1- it can improve \textbf{exploration} in sparse rewards environments, e.g. by computing an intrinsic reward based on the novelty of the states or the information gain; 2- it can push the agent to maximize its \textbf{empowerment} by rewarding the agent if it is heading towards areas where it controls its environment; 3- it can help the agent to learn a relevant \textbf{state representation}.
\item[Skill learning]: We define skill learning as the agent's ability to construct task-independent and reusable skills in an efficient way. There are two core components taking advantage of this motivation: one is about the ability of an agent to learn a \textbf{representation of diverse skills} in order to achieve them, the other one is about wisely choosing the skills to learn with a \textbf{curriculum}. Thereby, unlike previous classifications, we differentiate the motivation which builds the abstract meaning of a skill and the motivation which chooses the skill.
%learning to represent goals by solving policies which are associated, the other one is about wisely learning those skills with a curriculum. 
%self-generation of goals or skills, how to learn a skill, rec int for intra option, ability of an agent to learn a representation of diverse skills in an unsupervisedway
\end{description}

\begin{figure*}
\begin{centering}
\includegraphics[width=15cm]{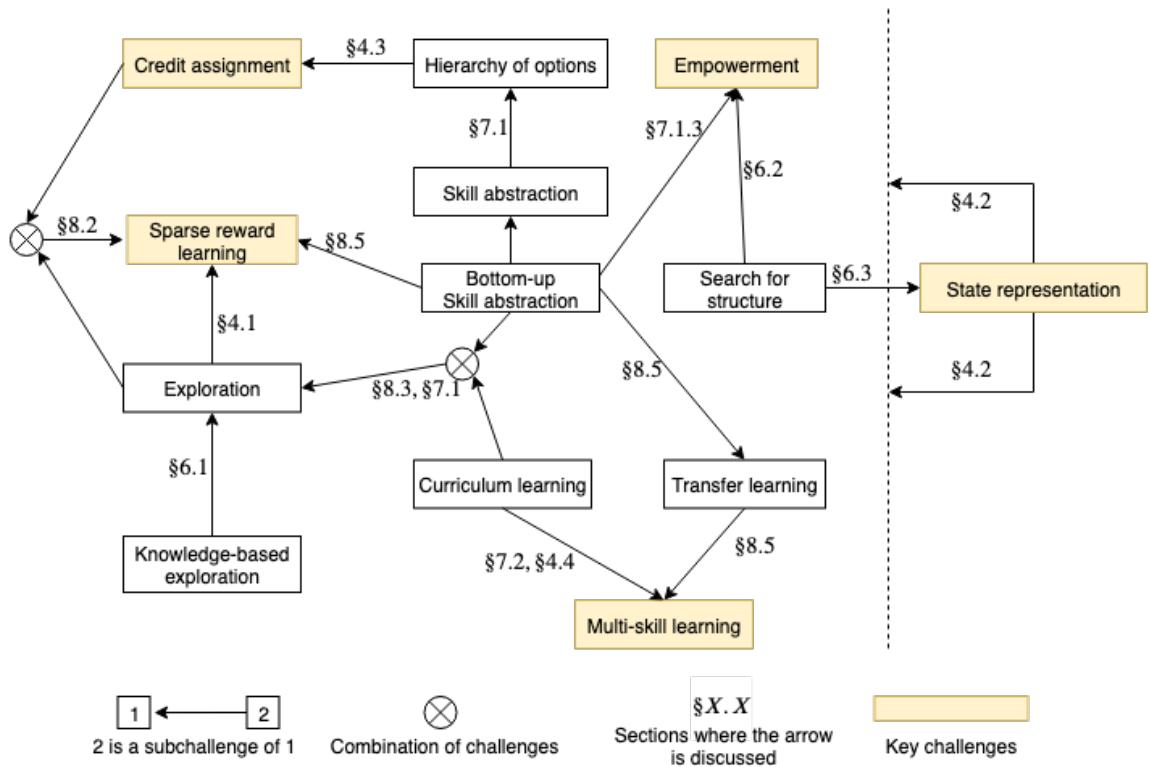}
\caption{Dependencies between challenges and their relation with our classification.}
\label{im:challenges}
\end{centering}
\end{figure*}

In figure \ref{im:challenges}, we summarize the relations between different challenges. For each relation, we refer to the appropriate section in which we discuss this relation.

Apart from our classification, some social motivations reward inequity or peace \cite{perolat2017multi,hughes2018inequity}. They deviate from the standard definition of an IM and are very specific to cooperative games. In this case, the reward is independent from the human, but still depends on the feedback of another agent. Therefore, we will not detail these works.

In the next two sections, we review the state-of-the-art by following the classification proposed in Table \ref{tab:tableofcontents}. 
\begin{table*}
\centering
\begin{minipage}{0.49\linewidth}
\begin{tabular}{|l|}
    \hline
    \hfill {\textbf{Knowledge acquisition}} \hfill\null  \\ 
    \hline
    \hline
    \textbf{Exploration}\\
    \hline
    Prediction error\\%lead the agent towards areas where the forward model is not accurate
    State novelty\\%add an intrinsic bonus when the agent goes into a state in which it usually never goes, count-based or pseudo-count based on density model
    Novelty as discrepancy towards other states\\%evaluate state novelty as the distance between a state and the states usually covered
    Information gain\\%eward based on the reduction of uncertainty on environment’s dynam-ics or learning progress
    \hline
    \hline
    \textbf{Empowerment}\\%maximize the empowerment, i.e. the agent is rewarded if it is heading towards areas where it controls its environment. 
    \hline
    \hline
    \textbf{Learning a relevant state representation}\\
    % Intrinsic  motivation enables to construct a policy generating a distribution of interactions (or sequence of states) allowing the construction of a disentangled feature space/to construct a feature space with meaningful properties
    \hline
    State space as a measure of distance\\
    One feature for one object of interaction\\
    \hline

\end{tabular}
\end{minipage}
\begin{minipage}{0.49\linewidth}
\begin{tabular}{|l|}
    \hline
    \hfill {\textbf{Skill learning}} \hfill\null  \\ 
    \hline
    \hline
    \textbf{Skill abstraction}\\
    %self-generation of goals or skills, how to learn a skill, rec int for intra option, ability of an agent to learn a representation of diverse skills in an unsupervisedway
    \hline
    Building the goal space from the state space \\%Using state space to generate goals and compute intrinsec rewards\\
    Mutual information between goals and trajectories\\
    %maximise mutual information between a goal and its trajectory
    \hline
    \hline
    \textbf{Curriculum learning}\\
    %rec int for inter option (intra option experte)
    \hline
    Goal sampling\\
    Multi-armed bandit\\
    Adversarial training\\
    \hline

\end{tabular}
\end{minipage}
 \caption{Classification of the use of IMs in DRL.}\label{tab:tableofcontents}

\end{table*}

\section{Knowledge acquisition}\label{motiv}

In this part, we survey the work related to knowledge acquisition according to the three challenges addressed by this approach. The more significant is undoubtedly the \textit{exploration problem} since it is the one which concentrates a large part of the literature. Although the amount of work is more timorous, we will see that \textit{state representation} can also take advantage of an active search for knowledge and that maximizing \textit{empowerment} is generally sufficient to produce interesting behaviors.
We focus our study on recent work and recommend to the interested reader to look at \cite{schmidhuber2010formal} for an overview on older methods on knowledge acquisition in RL.

%The second one is about acquiring skills, the agent tries to quickly develop reusable skills. In both case we will see it can be useful for tasks learning.

%Dans cet article, nous proposons de catégoriser les différentes motivations intrinsèques utilisées en AR en trois classes de méta-compétences: l'acquisition de connaissances, l'\textit{empowerment} et la génération d'objectifs.

%\cite{perolat2017multi} peace, soutenabilité etc..
%\cite{yu2015emotional} justice bien être
%\cite{hughes2018inequity}
%\cite{moerland2018emotion} survey emotions

\subsection{Exploration}\label{curiosity}

This subsection describes the three main methods tackling the \textit{exploration problem}. The first uses \textit{prediction error}, the second and third evaluate \textit{state novelty} and the fourth is based on \textit{information gain}. In each case, the IM completes an exploration policy.

\subsubsection{Prediction error}

Here, the idea is to lead the agent towards tbe areas where the prediction of the state following a state-action tuple is difficult. We can formalize this intrinsic reward by the \textit{prediction error} on the next state, computed as the distance between real and the predicted next state:
\begin{equation}
    R_{int}(s_t,s_{t+1})= {||g(s_{t+1})-F(g(s_t),a_t) ||}_2
\end{equation}

where $g$ is a generic function (e.g. identity or a learnt one) encoding the state space into a feature space and $F$ is a model of the environmental dynamics. In the following, we consider that $F$ is a neural network that learns a \textit{forward model} predicting the next encoded state given the current encoded state and action. We will see that learning a relevant function $g$ is the main challenge here.\\

The first natural idea to test is whether a function $g$ is required. \citename{burda2019largescale} learnt the forward model from the \textbf{ground state space} and observed it was inefficient when the state space is large. In fact, the L2\footnote{Euclidian distance.} distance is meaningless in such high-dimensional state space. In contrast, they raise up that \textbf{random features} extracted from a random neural network can be very competitive with other state-of-art methods. However they poorly generalize to environment changes. An other model, \textbf{Dynamic Auto-Encoder (Dynamic-AE)} \cite{stadie2015incentivizing}, computes the distance between the predicted and the real state in a state space compressed with an auto-encoder \cite{hinton2006reducing}. $g$ is then the encoding part of the auto-encoder. However this approach only slightly improves the results over Boltzmann exploration on some standard Atari games.

These methods are unable to handle the local stochasticity of the environment \cite{burda2019largescale}. For example, it turns out that adding random noise in a 3D environment attracts the agent; it will passively watch the noise since it will be unable to predict the next observation. \label{tele} This problem is also called \textit{the white-noise} problem \cite{pathak2017curiosity,schmidhuber2010formal}.\\% A potential solution would be to make sure that transitions can be learnt, \textit{i.e.} that transitions are not too stochastic, %but this problem is difficult to solve in practice \cite{lopes2012exploration}. 

To tackle exploration with local stochasticity, the \textbf{intrinsic curiosity module (ICM)} \cite{pathak2017curiosity} learns a state representation function $g$ end-to-end with an \textit{inverse model} (i.e. a model which predicts the action done between two states). Thus, the function $g$ is constrained to represent things that can be controlled by the agent. Secondly, the forward model $F$ used in ICM predicts, in the feature space computed by $g$, the next state given the action and the current state. The prediction error does not incorporate the white-noise that does not depend on actions, so it will not be represented in the feature state space. ICM notably allows the agent to explore its environment in the games \textit{VizDoom} et \textit{Super Mario Bros}. In \textit{Super Mario Bros}, the agent crosses 30\% of the first level without extrinsic reward. However one major drawback is the incapacity of the agent to keep in its representation what depends on its long-term control, for example, it may perceive the consequences of its actions several steps later. Building a similar action space, but with other improvements, \textbf{Exploration with Mutual Information (EMI)} \cite{pmlr-v97-kim19a} significantly outperforms previous works on Atari but at the cost of several complex layers. EMI transfers the complexity of learning a forward model into the learning of a space and action representation through the maximization of $I([s,a];s')$ and $I([s,s'];a)$. Then, the forward model $F$ is constrained to be a simple linear model in the representation space. Some transitions remain strongly non-linear and hard to predict (such as a screen change); while it acts like noise for previous approaches, EMI introduces a \textit{model error} which offload the linear model from these type of errors. EMI avoids the white-noise problem and proves to be able to construct an embedding space related to positions. However, similarly to ICM, features depending on long-term control are not included in the representation.

\subsubsection{State novelty}\label{sec:novelty}

There is a large literature on the measure of the \textit{state novelty} as IM. At the beginning, the intuition was to add an intrinsic bonus when the agent goes into a state in which it usually never goes \cite{brafman2002r,kearns2002near}. These methods are said to be \textit{count-based}. As the agent visits a state, the intrinsic reward associated with this state decreases. It can be formalized with:
\begin{equation}
    R_{int}(s_t) = \frac{1}{N(s_t)}
\end{equation}
where $N(s)$ is the number of times that the state $s$ has been visited. Although this method is efficient in a tabular environment (with a discrete state space), it is hardly applicable when states are numerous or continuous since an agent never really returns in the same state.\\

A first solution proposed by \citename{tang2017exploration}, called \textbf{TRPO-AE-hash}, is to hash the state space when it is too large. However these results are only slightly better than those obtained with a classic exploration policy. Other attempts of adaptation to a very large state space have been proposed, like \textbf{DDQN-PC} \cite{bellemare2016unifying}, \textbf{A3C+} \cite{bellemare2016unifying} or \textbf{DQN-PixelCNN} \cite{ostrovski2017count}, which rely on density models \cite{van2016conditional,bellemare2014skip}. Density models allow to compute the \textit{pseudo-count} \cite{bellemare2016unifying}, which is an adaptation of the counting enabling its generalization from a state towards neighbourhood states. The intrinsic reward is then:
\begin{equation}
        R_{int}(s_t) = \frac{1}{\hat{N}(s_t)}
\end{equation}
where $\hat{N}(s_t)$ is the pseudo-count defined as:
\begin{equation}
        \hat{N}(s_t) = \frac{\rho(s)(1-\rho'(s))}{\rho'(s)-\rho(s)}
\end{equation}
with $\rho(s)$ the density model which outputs a probability of observing $s$, and $\rho'(s)$ the probability to observe $s$ after one more pass on $s$. \\

Although the algorithms based on density models work on environments with sparse rewards, density models add an important complexity layer \cite{ostrovski2017count}. Two approaches manage to preserve the quality of observed exploration while decreasing the computational complexity of the pseudo-count:
\begin{itemize}
\item \textbf{$\phi$-EB} \cite{martin2017count} avoids modelling the density on the raw state space, but on a feature space induced by the computation of $V(s)$. The results are impressive on Montezuma's revenge considering the cutback in the computational cost. The latent space can also be computed with a variational auto-encoder \cite{vezzani2019learning}.
\item \textbf{DQN+SR} \cite{machado2018count} considers the norm of the successor representation \cite{kulkarni2016deep} as an intrinsic reward. To justify this choice, the authors explain that this bonus is correlated to the counting. A slight counterpart is that their agent needs to train a forward model as an auxiliary task to learn a representation without extrinsic rewards.
\end{itemize}

Finally, \textbf{RND} \cite{burda2018exploration} implicitly does a computation roughly similar to pseudo-count, with a lower complexity and better final results. They assess \textit{state novelty} by distilling a random neural network (with fix weights) into an other neural network. For every state, the random network produces random features which are continuous. The second network learns to reproduce the output of the random network for each state. The prediction error is the reward. it is like rewarding \textit{state novelty} since the error will be high when the second network has still not visited many times the considered state, and the error will be low after several visits. RND holds the highest score on Montezuma's revenge, but with a significantly larger number of steps (see Table \ref{tab:curiosity}). On a short timescale, this method is outperformed by previous approaches \cite{machado2018count}. Lastly, random features can be insufficient to represent the wealth of an environment.

\subsubsection{Novelty as discrepancy towards other states}\label{sec:discrepancy}

An other way to evaluate the \textit{state novelty} is to estimate the distance between a state and the usually covered states. With $D$ as a distance function and $B$ as a distribution of states among a moving buffer, we can describe this kind of reward as :
\begin{equation}
    R_{int}(s_t) = \mathbb{E}_{s' \sim B}\left[D(s_t;s')\right].
\end{equation}

Among methods based on the computation of \textit{state novelty}, \citename{stanton2018deep} distinguish \textit{inter-episodes novelty}, and \textit{intra-episodes novelty}. Both have different properties, inter-episode novelty learns an exploration policy across episodes, while intra-episode novelty seeks for a policy which optimally explores inside an episode. Typically, intra-episodes novelty will reset the state count at the beginning of each episode. It can be particularly adequate when the environment is not strictly identical at each episode; in this case the agent must learn how to explore rather than going into states in which it has never been. Previous sections only aaddressed inter-episodes novelty (Section \ref{sec:novelty}); here we distinguish both.\\

First we will focus on methods using intra-episode novelty, and then we will report methods maximizing inter-episodes novelty.

\paragraph{Intra-episode novelty:} \textbf{Informed exploration} \cite{oh2015action} uses a forward model to predict which action will bring the agent in the most different states compared to its $d$ last visited states. The authors use a Gaussian kernel as a metric. However they do not use this distance as an intrinsic reward but as a way to choose the action instead of $\epsilon$-greedy strategy; therefore, their agent explores only one step ahead. They slightly improve standard exploration methods on some Atari games. At the opposite, methods which sum extrinsic and intrinsic rewards can take advantage of the long-term maximization of the intrinsic reward. The \textbf{episodic curiosity module (ECO)} \cite{savinov2018episodic} deepens the idea of intra-option novelty by taking inspiration from episodic memory. The proposed model contains a comparison module (trained with a siamese architecture \cite{ZagoruykoK15}) able to give a bonus if the agent is far from the states contained in a buffer. Therefore, it computes the probability that the number of necessary actions to go to a selected state (in a buffer) from the current state is below a threshold. By storing sparse states into a buffer, the agent sets reference points in the environment, as if it is partitioning the environment, and tries to get away from them. The probability that the agent is away from every buffer is used as an intrinsic reward. This model is suited for 3D environments like \textit{DMLab} \cite{beattie2016deepmind} or \textit{VizDoom} \cite{kempka2016vizdoom} and enables an agent to explore the overall environment, even if its structure changes at each episode. However, to compute the intrinsic reward, the agent has to compare its current observation to each memorized state. Scaling up this method may then be difficult when the state space is rich since it will require more states to efficiently partition the state space. On the other side, this method does not suffer from the white-noise problem (cf. \S \ref{tele}).

\paragraph{Inter-episodes novelty:}%From now on, we will focus again an inter-episodes novelty. 
\textbf{EX$^2$} \cite{fu2017ex2} is a popular approach that learns a discriminator to differentiate states from each other: when the discriminator does not manage to differentiate the current state from those in the buffer, it means that the agent has not visited this state enough and it will be rewarded, and inversely if it is able to make the differentiation. Here, the comparison between states is sampled from a buffer, implying the necessity to have a large buffer. To avoid this, it is possible to distill recent states into a distribution. Doing so, \textbf{CB} \cite{kim2019curiosity} mixes up \textit{prediction error} and \textit{state novelty}. It gets inspiration from the deep variational information bottleneck \cite{alemi2016deep}: it computes a latent state space by maximizing the mutual information between the state value and this latent space, with a latent distribution as entropic as possible. The intrinsic reward for a state is then the KL-divergence between a fixed diagonal Gaussian prior and the posterior of the distribution of latent variables. It results that, as long as the agent does not find any reward, it will look for rare states which have a distribution in the latent space that is different from the prior (common states fit the prior). When the agent finds the reward, the latent distribution will be different from the prior and the intrinsic reward will guide the agent towards interesting areas. While this approach provides good results on Gravitar and Solaris, it requires an extrinsic reward to avoid the white-noise problem. A similar KL-divergence intrinsic reward can be found in \textbf{VSIMR} \cite{klissarovvariational}, but with a standard variational auto-encoder (VAE).

Finally, \textbf{state marginal matching (SMM)} \cite{lee2019efficient} is a method close to \textit{pseudo-count}, but it recomputes the KL-divergence between state distribution induced by the policy and a target distribution. In fact, when the target distribution is uniform, the agent strives to maximize the state entropy. It explicitly encourages the agent to uniformly go into all states of the environment. This objective is also combined with the discriminative policy objective which induces a distribution of trajectories. This second objective is explained in Section \ref{miskill} and its benefits for exploration are further discussed in Section \ref{sec:curriculum}. The results of this approach are interesting on navigation tasks, but they are not compared to usual exploration benchmarks.

%do not compare them-self on usual exploration benchmarks.\\
 %It could be a way to overcome the RND \cite{burda2018exploration} issue to handle long-term exploration. %Dans \textit{Montezuma's revenge}, l'agent utilise ses clés pour ouvrir les premières portes qu'il voit et n'arrive donc pas à accéder aux deux dernières portes.

\subsubsection{Information gain}\label{sec:info_gain}

The information gain is a reward based on the reduction of uncertainty on the environment dynamics \cite{oudeyer2009intrinsic,little2013learning}, which can also be assimilated to Bayesian surprise \cite{itti2006bayesian,schmidhuber2008driven}. It also has common properties with learning progress \cite{oudeyer2009intrinsic,schmidhuber1991curious,frank2014curiosity}, which is the improvement of the agent performance on its task. This allows, on one side, to push the agent towards areas it does not know, and on the other side to prevent attraction towards stochastic areas. Indeed, if the area is deterministic, environment transitions are predictable and the uncertainty about its dynamics can decrease. At the opposite, if transitions are stochastic, the agent turns out to be unable to predict transitions and does not reduce uncertainty. It results that these methods efficiently tackle the white-noise problem. If $\theta$ is the parameter set of a dynamic parametric model and $U$ refers to uncertainty, this can be defined as:
\begin{equation}
    R_{int}(s_t,s_{t+k}) =  U_{t+k}(\theta) - U_t(\theta).
\end{equation}

The exploration strategy \textbf{VIME} \cite{houthooft2016vime} formalizes learning progress in a Bayesian way. The interest of Bayesian approaches is to be able to measure the uncertainty on the learned model \cite{blundell2015weight}. Thus, the agent approximates these dynamics with a Bayesian neural network \cite{graves2011practical}, and computes the reward as the uncertainty reduction on weights. In other words, the agent tries to do actions which are informative on the dynamics. However, the interest of the proposed algorithm is shown only on simple environments and the reward can be computationally expensive to compute. \citename{achiam2017surprise} proposed a similar method, with comparable results, using deterministic neural networks, which are simpler and quicker to apply. The weak performance of both models is probably due to the difficulty to retrieve the uncertainty reduction by rigorously following the mathematical formalism of information gain.

Therefore, more creatively, \citename{pathak2019self} train several (generally five) forward models in a feature space and estimate their mean predictions. The more the models are trained on a state-action tuple, the more they will converge to the expectation value of the next state features. The intrinsic reward is then the variance of the ensemble of predictions. The benefits are that the variance is high when forward models are not learned, but low when the noise comes from the environment since all the models will converge to the mean value. It appears that this method is competitive with state of the art approaches \cite{burda2019largescale}. However the main intrinsic issue is computational since it requires multiple forward models to train. A similar intuition can be found in \textbf{JDRX} \cite{shyam2018model} with equations derived from the information gain. They mostly differ from \citename{pathak2019self} by using Jensen-Shannon divergence between distributions of several stochastic forward models instead of the variance across outputs of deterministic models. One potential drawback is that the model relies on parametric distributions on the state space.
%The authors also propose a variant, \textbf{MAX}, which use exploration bonus to find a plan, making the algorithm more sample efficient. One potential drawback is that, the model uses, and relies, on parametric distributions on the state space.%In a similar way, \citename{achiam2017surprise} replace the Bayesian model by a classic neural network followed by a factorized Gaussian probability distribution. Two rewards are evaluated: the first one (\textbf{NLL}) uses as intrinsic bonus the cross entropy of the prediction, and the second one (\textbf{AKL}) the improvement of the prediction between the time $t$ and after $k$ improvements at $t+k$. Although these methods are simpler than VIME, they do not improve over VIME.
\begin{table*}
\centering
\begin{threeparttable}
\begin{tabular}{|l|l|l|l|l|}
  \hline
   Method & Stoch \footnotemark[4] & Computational cost & Score & Steps \\
  \hline
  \hline
    \textbf{Prediction error}  & & & &  \\ 
    \hline
    No features \cite{burda2019largescale} & No & HD \footnotemark[1] FM\footnotemark[5] & $\sim 160$ & 200M \\
    & & VAE & & \\
    \hline
    Dynamic-AE \cite{stadie2015incentivizing} & No & FM / AE & $0$ & 5M \\
    \hline
    Random features \cite{burda2019largescale} & No & FM & $\sim 250$ & 200M \\
    \hline
    VAE features \cite{burda2019largescale} & No & FM / VAE & $\sim 450$ & 200M \\
    \hline
    ICM features \cite{burda2019largescale} & Yes & Inverse model & $\sim 160$ & 200M \\
    \cite{pmlr-v97-kim19a} &  & FM & $161$   & 50M \\
    %\hline
    %AR4E \cite{ohlearning} & Yes & Inverse model & n/a & n/a \\
    %& & HD forward model & & \\
    \hline
    EMI \cite{pmlr-v97-kim19a} & Yes & Large architecture & $387$ & 40M \\ 
    &  & Error model & &  \\ 
    \hline
    \hline
    \textbf{State novelty} & & & & \\
    \hline
    \hline
    TRPO-AE-hash \cite{tang2017exploration} & Partially & SimHash / AE & $75$ & 50M  \\ 
    \hline
    DDQN-PC \cite{tang2017exploration} & Partially & CTS & $3459$ & 100M  \\ 
    \hline
    DQN-PixelCNN \cite{ostrovski2017count}& Partially & PixelCNN & $\sim 1670$ & 100M  \\ 
    \hline
    $\phi$-EB \cite{martin2017count}& Partially & LD \footnotemark[2] density model & $2745$ & 100M \\
    \hline
    DQN+SR \cite{machado2018count}& Yes & Successor features & $1778$ & 100M \\
    & & HD FM & & \\   

    \hline
    %DORA \cite{choshen2018dora}& Yes & Other MDP & n/a & n/a \\
    %\hline
    RND \cite{burda2018exploration} & Partially & One learning NN\footnotemark[3] & $8152$ & 1970M \\
    \cite{machado2018count}& & One non-learning NN & $524$ & 100M \\
    \cite{pmlr-v97-kim19a} & & & $377$ & 50M \\
    \hline 
    Informed exploration \cite{oh2015action} & No & FM & n/a & n/a \\
    \hline
    EX$^2$ \cite{fu2017ex2} & Partially & Discriminator & n/a & n/a \\
    \cite{pmlr-v97-kim19a} & & & 0 & 50M \\
    \hline 
    CB \cite{kim2019curiosity}& No & IB & $\sim 1700$ & n/a \\
    \hline
    VSIMR \cite{klissarovvariational}& No & VAE & n/a & n/a \\
    \hline
    ECO \cite{savinov2018episodic} & Yes & Siamese architecture & n/a & n/a \\
    & & Several Comparisons & &  \\
    \hline
    SMM \cite{lee2019efficient} & ~No & VAE & n/a & n/a \\
    & & Discriminator & & \\
    \hline
    \hline
    \textbf{Information gain} & & & & \\
    \hline
    \hline
    VIME \cite{houthooft2016vime} & Yes & Bayesian FM & n/a & n/a \\
    %& & Intrinsic reward & & \\
    \hline
    AKL \cite{achiam2017surprise} & Yes & Stochastic FM & n/a & n/a \\
    \hline
    Ensembles \cite{pathak2019self} & Yes & 5 LD FM & n/a & n/a \\
    \hline
    MAX \cite{shyam2018model} & Yes & 3 Stochastic FM & n/a & n/a \\
    \hline
    
\end{tabular}
  \begin{tablenotes}
    \item[1] High-dimensional.
    \item[2] Low-dimensional.
    \item[3] Neural network.
    \item[4] Stochasticity.
    \item[5] Forward model.

  \end{tablenotes}
\end{threeparttable}

 \caption{Comparison between exploration strategies with IM. Stochasticity indicates whether the model handles the white-noise problem (a deeper analysis is provided in \S\ref{sec:stochasticity}). Computational cost refers to highly expensive models added to standard RL algorithm. We also display the mean score on \textit{Montezuma's revenge} (Score) and the number of timesteps executed to achieve this score (Steps). We also integrate results of some methods tested in other papers than the original one. Our table does not pretend to be an exhaustive comparison of methods but tries to give an intuition on their relative advantages. We invite the reader to have a look at the original articles for a more thorough study.}\label{tab:curiosity}
\end{table*}

%\footnotetext[1]{High-dimensional}
%\footnotetext[2]{Low-dimensional}
%\footnotetext[3]{Neural network}

\subsubsection{Conclusion}

To conclude, the exploration problem is probably the largest use case for IM. We provide a synthesis of our analysis in Table \ref{tab:curiosity}. A complementary benchmark can be found in \cite{taiga2019benchmarking}. There are multiple distinct heads: most count-based approaches are adapted for fully-observable MDPs, like \textit{Montezuma's revenge}; error prediction is relatively simple but relies on a good state representation; information gain methods are particularly adequate to prevent stochasticity to interfere with exploration but are harder to compute. In fact, before choosing the right exploration method, it is important to consider the trade-off between computational cost and efficiency. On simple environments, simple methods can perform well. So far, the more complex tested environment is Montezuma's environment, however it might be necessary to consider larger/infinite environments like Minecraft \cite{johnson2016malmo} to wisely advice and compare these methods. Indeed, it would be important to know how count-based methods \cite{ostrovski2017count} or ECO \cite{savinov2018episodic} scale to these kind of environments. 

Furthermore, some works tackling exploration through a curriculum over skills are described in Section \ref{sec:curriculum}.
%How will count-based methods \cite{ostrovski2017count} or EC \cite{savinov2018episodic} scale to these environments ? 
Lastly, to our knowledge, few works tried to adapt these exploration processes to a multi-agent scenario, which is known to have an exponentially larger state space \cite{oliehoek2012decentralized}. Among them, \citename{iqbal2019coordinated} introduce different ways to guide the exploration process, but only consider very simple tabular environments.

\subsection{\textit{Empowerment}} \label{empowerment}

As presented in Section \ref{sect:background_empowerment}, an agent that maximizes empowerment tries to have the most control on its environment. To maximize empowerment in RL, the agent is rewarded if it is heading towards areas where it controls its environment. The intrinsic reward function is then defined as:
\begin{align}
    R_{int}(s,a,s') &= \Sigma(s') \nonumber \\
     & \approx -\mathbb{E}_{\omega (a|s)} \log \omega (a|s) + \mathbb{E}_{p(s'|a,s)\omega (a|s)}\log p(a|s,s') \label{eq:entropy2}. 
\end{align}

where $\omega (a|s)$ is the distribution choosing actions $a_t^n$. Ideally, $\omega (a|s)$ is the distribution maximizing Equation \eqref{eq:entropy2} in accordance with Equation \eqref{eq:meaning}.

The problem is that $p(a|s,s')$ is hard to obtain because it requires $p(s'|a,s)$ which is intractable. \\

\citename{mohamed2015variational} propose to compute the empowerment by approximating Equation \eqref{eq:entropy2}. To do this, they compute a lower bound of mutual information, used in many other works (Section \ref{miskill}):
\begin{equation}
    I(a;s'|s) \geq H(a|s) + \mathbb{E}_{p(s'|a,s)\omega (a|s)}\log q_{\xi}(a|s,s'). \label{eq:vlb}
\end{equation}

The idea is to learn an approximator $q_{\xi}$ of the probability distribution $p(a|s,s')$ in a supervised way with maximum likelihood method by using data received by the agent from its environment. This approach allows to generalize the computation of empowerment in order to process continuous observations. In this work, experiments show that the maximization of \textit{empowerment} is particularly useful in dynamic environments, i.e. environments where the agent's state can change even if the executed action is stationary (e.g. the agent does not move). The classic example provided in \citename{mohamed2015variational} is the prey-predator environment: the prey is the learner and tries to avoid to be caught as its death will cause a loss of control on the next states. Implicitly, the prey avoids to die by maximizing its \textit{empowerment}. In contrast to a dynamic environment, a static environment has a static optimal policy (the agent stops moving when it finds the best state) making \textit{empowerment} as an intrinsic reward less interesting according to a task. However, experiments proposed in \citename{mohamed2015variational} use planning methods to estimate \textit{empowerment} instead of interactions with the environment to collect data, which implies the use of a forward model.\\

\textbf{VIC} \cite{gregor2016variational} tries to maximize \textit{empowerment} with interactions with the environment using $\omega(a|s) = \pi(a|s)$. The intrinsic reward then becomes :
\begin{equation}
R_{int}(a,h) = -\log \pi(a|h) + \log \pi(a|s',h)
\end{equation}
where $h$ is the observation history (including current observation and action). The experiments on diverse environments show that learned trajectories lead to diverse areas and that a pretraining using \textit{empowerment} helps to learn a task. However, learned tasks are still relatively simple. The main issue may be that the \textit{empowerment} is hard to compute. We found few works related to \textit{empowerment} not following the formalism, while still rewarding the control of the agent.

Instead of directly using mutual information, \textbf{Mega-reward} \cite{song2019mega} cuts out the pixel space into a matrix which defines the probability of control of the corresponded part of the image. The intrinsic reward is then the matrix sum. They also show that the matrix can act as a mask to hide uncontrollable features, what other intrinsic exploration methods \cite{burda2018exploration} can benefit from to reduce the white-noise problem in a long-term way (as opposite to ICM method which detects short-term controllable features). However the method is inherently linked to pixel state environments. \citename{chuck2019hypothesis} provide a specific architecture relying on multiple assumptions such as the fact that an object can not spontaneously change its direction or its proximity to objects it interacts with. The agent formulates hypothesis on the \textit{controllability} of objects, which it tries to verify through a specific policy rewarded with an intrinsic verification process. Checked hypothesis can then be used directly as skills.\\

\textit{Empowerment} may also be interesting in multi-agents RL. Multi-agents RL is similar to mono-agent RL except that several agent learn simultaneously to solve a task and have to coordinate with each other. \citename{jaques2019social} show that in a non-cooperative game, as social dilemma \cite{leibo2017multi}, an \textit{empowerment}-based intrinsic reward could stabilize the learning process; the agent acts in order to influence other agents instead of looking for extrinsically rewarded behaviors. In fact, it compensates for the decrease of individual reward caused by a policy maximizing the long-term reward of all the agents.

To sum up, \textit{empowerment} is an interesting method to avoid an extrinsic reward and keep various complex behaviors. The main difficulty using \textit{empowerment} in RL is its complexity. Several approaches use an environment model to compute the reward based on \textit{empowerment} \cite{mohamed2015variational,de2018unified}. However the very essence of RL is that the agent does not know \textit{a priori} environment dynamics or the reward function. Existing work in this context remains relatively limited and is not sufficient to demonstrate the potential of \textit{empowerment} to help the learning process. It is interesting to note that \textit{empowerment} can push an agent to learn behaviors even in \textit{a priori} static environments. Indeed, let us assume that the agent does not choose primitive actions directly, but {options} instead. If it has not learned options, it will be unable to distinguish them, thus it is as if the agent had no control on the environment. On the contrary, if its options are perfectly distinguishable in the state space, the agent has control on its environment. In fact, the issue is not about choosing the states maximizing {empowerment}, but about defining options which increase overall \textit{empowerment}. We will come back to this point in Section \ref{gen_goal}.

\subsection{Learning a relevant state representation}\label{sec:staterep}

Learning a relevant state representation is the ability of the agent to project its raw state onto a feature space with meaningful properties (cf. \S\ref{sec:staterepresentation}). Random policies as well as task-specific policies only access a subset of the state space, which can prevent the construction of a disjoint state representation. Indeed the distribution and sequence of states reached by the agent strongly depends on the overall policy of the agent. IM brings here high interests as it enables to construct a policy generating the right distribution of interactions.

Generally, two successive states must be close in the built feature space. Taking into account states independently is not sufficient to produce an efficient representation. Moreover, it is desirable to separate the different objects to which the agent can pay attention since it facilitates the learning process. We will study in these subsection how IM gives a valuable alternative to standard methods, by providing interactions in the environment that take into account knowledge on causal links between observations \cite{caselles2019symmetry,de2018integrating}. An other interest is the fact that no supervision is needed.

\subsubsection{State space as a measure of distance}\label{sec:statemeasure}
%remettre en question la L2 distance dans le labyrinth

The first set of work tries to fit distance in the representation space with distance in terms of action in the state space. Two methods propose a specific reward to learn a state representation for which the L2 distance between two states is proportional to the minimal number of actions required to connect these two states. % needed to go from one state to the other. 
%\citename{venkattaramanujam2019self} try to learn an embedding space where the L2 distance between two states is proportional to the number of action needed to go from one state to the other. 
To do so, \citename{venkattaramanujam2019self} train a predictor of state distance on states separated by random actions. However, as admitted by the authors, this is not strictly a distance since the distance between a state with itself can be non-zero (the agent does a loop). In addition, they need the ability to reset the environment in all states. Instead of relying on random actions, \citename{florensa2019self} consider a goal-parameterized problem where the goal space is the state space. The intrinsic reward function is composed of two parts: the first part imposes that the agent reaches the goal with a binary reward, the second part constrains the distance between two consecutive steps to be around 0. However, they assume that a goal is provided in the state space and they lack more elaborated experiments showing the relevance of their approach. 

\citename{ghosh2018learning} also assume that a goal-conditioned policy trained with IM is available (where the goal space is the state space). Then they use trajectories of this policy to learn a state space representation where the L2 distance between two goal states corresponds to the expected KL-divergence of policies from an uniform distribution of states. Interestingly, they manage to heavily differentiate subsets of the state space which are separated by a bottleneck.

\subsubsection{One feature for one object of interaction}

\citename{thomas2017independently} try to learn independent factors of variation in the embedding space. The goal is presented as a variation of one feature in the embedded space, which is learnt simultaneously with the policy. For example, such feature can be the light of a room, and a policy relative to this factor of variation can be the fact of switching it off or on. The reward is thus the maximization of the chosen variation factors in comparison with other variation factors. The agent manages to assimilate a factor of variation only to a deterministic static object but it is not clear how the agent can generalize across moving objects. This approach has been further extended to also represent factors of variation of uncontrollable features (an unalterable barrier for example) \cite{sawada2018disentangling}. 

\subsubsection{Conclusion}

To conclude, although most of the work does not consider the learning of state representation as a task in itself \cite{de2018integrating}, it allows to construct a state space with meaningful properties. We strongly believe that an active learning process is required to understand properties of the world. Interesting events exhibiting these properties are rare with random actions whereas it can be common with specific goals. Typically, it is easier for an agent to distinguish two different objects if it tries to move them independently. It will take much longer if the agent just waits for one movement to accidentally happen. As an other example, it can only understand the concept of distance by moving towards objects.

\section{Skill learning}\label{skill_learning}

In our everyday life, nobody has to think about having to move his arms' muscles to grasp an object. A command to take the object is just issued. This can be done because an acquired skill can be effortlessly reused. At the same time, while we learnt to grasp objects, we did not try to learn to move our ears because it is almost impossible. IM provides a useful tool to get learnable skills (or options) without the need of hand-engineering tasks. In this section, we will first review how an agent can learn a representation of various skills by using intrinsic rewards to learn intra-option policies. Secondly, we will present methods addressing how to choose which skills to train on, i.e. how to use IM to learn inter-option policies.

%first review how to learn a skill and then how to choose which skill to train on.

\subsection{Skill abstraction} \label{gen_goal}

Skill abstraction is the ability of an agent to learn a representation of diverse skills. Skills or goals generated by the agent are {options} (cf. \S\ref{sect:defis}). In comparison with multi-objective RL \cite{liu2015multiobjective}, skills are here learned in an unsupervised way and the agent generally learns on two timescales: on the one hand, it generates options and learns associated intra-options policies using an intrinsic reward; on the other hand, if a global objective (or task) exists, it will learn to use its skills to realize this global objective using the extrinsic reward associated to the task. To learn intra-options policies, it is possible to use UVFA or HER (cf. \S \ref{uvfa}) since the reward function $R(s,a,s',g')$ can be computed without additional interactions when we only use an intrinsic reward.%global objective = task

Key aspects are first to learn interesting skills which can be transferred between several tasks. These skills can be even more transferable if they are uncorrelated from the learned task \cite{heess2016learning}. Secondly, temporal abstraction of executed actions through acquired skills makes the learning process easier. Let us take, as an example, MuJoCo \cite{todorov2012mujoco}, which is a commonly used environment in works related to skills. In this environment, the joints of the robot can be controlled by an agent to achieve, for example, locomotion tasks. The idea of some works we will study is to generate skills like \textit{move forward} or \textit{move backward} with an intrinsic reward. These skills can then be used for a navigation task.

In the following, we will present several works incorporating an intrinsic reward based on expert knowledge in a hierarchical algorithm, demonstrating the potential of the approach. Then we will study two main research directions on self-generation of goals. The first one uses the state space to generate goals and compute the intrinsic reward; the second one uses information theory to generate skills based on a diversity heuristic.

\subsubsection{Intrinsic rewards with expert knowledge}

In this part, we will first study an article highlighting the promises of the approach, but relying on strong assumptions. Then we will describe some used heuristics which can not generalize to all environments.

\paragraph{Strong assumptions:}Seminal work shows the interest of decomposing hierarchically actions. Among them, \citename{kulkarni2016hierarchical} present the \textbf{hierarchical-DQN} in which the goal representation is expertly defined with tuples $(entity1,relation,entity2)$. An entity can be an object on the screen or an agent, and the relation notably refers to a distance. Therefore, the goal can be for the agent to reach an object. This reward is one if the goal is reached, zero otherwise. They show that it can help the learning process particularly when rewards are sparse like in \textit{Montezuma's revenge}. In fact, the more hierarchical the task is, the more required a hierarchical policy is \cite{complexity_exploration}.  However, by avoiding learning skill representation, \citename{kulkarni2016hierarchical} obfuscate the main problem: it is difficult to choose which features are interesting enough to be considered as goals in a large state space. 

\paragraph{Particular heuristics:} Other works demonstrate the potential of the approach using auxiliary objectives specific to the task \cite{riedmiller2018learning} or more abstract ones \cite{dilokthanakul2019feature,rafati2019unsupervised}. More particularly, an heuristic regularly used to generate skills is the search for the states acting as a bottleneck \cite{mcgovern2001automatic,menache2002q}. The main idea is to identify pivotal states relatively to the next visited states (e.g. a door). Recent works \cite{zhang2019scheduled,tomar2018successor} use successor representation \cite{kulkarni2016deep} to generalize the approach to continuous state space. Other heuristic can be the search for salient events \cite{chentanez2005intrinsically} such as changes in light. 

The limitation of this kind of works is that rewards are not sufficiently general to be applied in all environments. For example, there is no bottleneck state in an empty room whereas interesting skills can still be learned (going to the upper left corner).

\subsubsection{Building the goal space from the state space}\label{sec:goal_state}

Several works use the state space to construct a goal space, so as to consider every state as a potential goal. A distance between a goal state and the current state serves as an intrinsic reward. The goal state is chosen by an inter-option policy. Formally, we obtain : 
%Formally, when the last state of the option is considered, these approaches compute the intrinsic reward with:
\begin{equation}
    R_{int}(s_t,g_t)=D(f(s_t);f(g_t))
    \label{eq:distance_reward}
\end{equation}
where $D$ is a distance function, $g_t$ is the chosen goal,
$s_t$ is the current state and $f$ is a representation function which can be identity. When the direction is taken as intrinsic reward, it can be described with:
\begin{equation}
    R_{int}(s_t,g_t)=D(f(s_t)-f(s_f); g_t)
    \label{eq:direction_equation}
\end{equation}
where $s_f$ is the agent's state at the end of the option. In the following, we will describe methods emphasizing the complexity resulting of using identity function as $f$. As a second step, we will study different ways to define the function $f$ and how we can capitalize on it.

\paragraph{Ground goal space:} \textbf{Hierarchical Actor-Critic (HAC)} \cite{levy2018hierarchical} directly uses the state space as a goal space to learn three levels of option (the options from the second level are selected to fulfill the chosen option from the third level). A reward is given when the distance between states and goals (the same distance as in Equation \ref{eq:distance_reward}) is below a threshold. They take advantage of HER to avoid to directly use the threshold. Related to this work, \textbf{HIRO} \cite{nachum2019data} uses as a goal the difference between the initial state and the state at the end of the option (Equation \ref{eq:direction_equation}). The intrinsic reward allows to guide skills towards specific spatial areas. However, there are two problems in using the state space in the reward function. On the one hand, a distance (like L2) makes little sense in a very large space like an image composed of pixels. On the other hand, it is difficult to make an inter-option policy learn on a too large action space. Typically, an algorithm having as goals images can imply an action space of $84\times 84\times 3$ dimensions for the inter-option policy (in the case of an image with standard shape). Such a wide space is currently intractable, so these algorithms can only work on low-dimensional state spaces.

As a consequence, one key problem here is to build an efficient state space representation acting as a goal space  \cite{schwenker2019artificial}, i.e. to choose information that should not be lost during the compression of states into a new representation. Thus, the distance between two objectives would have a meaning and would be a good intrinsic reward.

\paragraph{Learning a state/goal representation:}
\textbf{FuN} \cite{vezhnevets2017feudal} uses as an intrinsic reward the direction taken in a feature space. The features used are those which are useful to the task, they are therefore inherently dependent of the extrinsic reward. It results that the agent can only learn skills related to the task and needs an access to the extrinsic reward. Then, \textbf{RIG} \cite{nair2018visual} proposes to build the feature space independently from the task with variational auto-encoder (VAE); but this approach can be very sensitive to distractors (i.e. useless features for the task or goal, inside states) and does not allow to correctly weight features. \textbf{Skew-fit} \cite{pong2019skew} recently improves over RIG by weighting rare states, leading to more diverse policies. Both RIG and Skew-fit are efficient in reaching a given state. \textbf{SFA-GWR-HRL} \cite{zhou2019vision} uses unsupervised methods like the algorithms of \textit{slow features analysis} \cite{wiskott2002slow} and \textit{growing when required} \cite{marsland2002self} to build a topological map. A hierarchical agent then uses nodes of the map, representing positions in the world, as a goal space. As opposed to other works, they manage to perform a navigation task through first-view visual inputs; but they do not compare their contribution to previous approaches. 

The current state-of-art method, \textbf{Sub-optimal representation learning} \cite{nachum2019near}, bounds sub-optimality of the goal representation with respect to the task, giving theoretical guarantees; the agent can solve the task with a hierarchical policy as precisely as with a simple low-level policy. The agent turns out to be able to learn a representation based on coordinates in a complex navigation task while the intra-option policy is agnostic from the task reward. Even though their results are impressive using low-scale top-view images as input, it would be interesting to test it on even larger/different state space.

%However, like FuN \cite{vezhnevets2017feudal}, the agent requires a dense reward.

\subsubsection{Mutual information between goals and trajectories} \label{miskill}

The second approach does not need a distance function but rather consists in maximizing mutual information between a goal and its associated trajectory. With $\tau$ the trajectory during the option, $s_i$ the initial state, $f$ a function selecting a part of the trajectory, $g_t$ a goal provided by an inter-option policy or sampled, we can compute the intrinsic reward as:
\begin{equation}
    R_{int}(s_t,g_t)=I(g_t,f(\tau)|s_i).
    \label{eq:im}
\end{equation}

Informally, it is about learning skills according to the ability of the agent to discern them from the trajectory (i.e. covered states) of the intra-option policy. In other words, the agent goes towards areas for which it can guess the option it has chosen. It enforces the building of diverse policies. We have stated at \S \ref{empowerment} that the \textit{empowerment} of an agent improves as skills are being distinguished. Work presented here implicitly increases the \textit{empowerment} of an agent, from the option policy point of view. Indeed, it maintains a high entropy on goals and associates a direction in the state space to a goal. Therefore, if $o$ is an option, $H(o|s)$ is maximal if the probability distribution is uniform, and $H(o|s,s')$ decreases as the agent learns to differentiate between options. 

In the following, we will first detail seminal works on this method. We will then focus our attention on their limits and the corresponding works addressing them.

\paragraph{Preliminary methods:} \textbf{SNN4HRL} \cite{florensa2017stochastic} is the first to learn skills by maximizing equation \eqref{eq:im}. Each goal is uniformly generated, so maximizing this equation is like minimizing $H(g|f(\tau))$ (cf. Equation \eqref{information}) which is equivalent to maximizing the intrinsic reward $\log q(g|f(\tau))$ (cf. Equation \eqref{eq:vlb}). In order to compute the probability $q(g|f(\tau))$, the state space is discretized into partitions, making it possible to count the number of visits of each partition for the current objective $g$. $f(\tau)$ assigns next state of an interaction to its partition. With the count, the agent can compute the probability $q$ with a simple normalization. Then, once the agent has learned the skills, it is integrated in a hierarchical structure in which a manager, or inter-option policy, chooses the goals to accomplish. Let us notice that the goal space is here discrete. In this paper, the main issue is that the state space have to be reliably partitioned, and this is not always possible. \textbf{VALOR} \cite{achiam2018variational} and \textbf{DIAYN} \cite{eysenbach2018diversity} reflect the same idea, but differ from SNN4HRL firstly by using a neural network rather than a discretization to compute $\log q(g|f(\tau))$ and secondly in choosing $f$ as a part of the trajectory of the skill in the environment. With VALOR, the agent manages to learn up to 10 different skills and up to 100 by gradually increasing the number of goals with a curriculum \cite{achiam2018variational}. For example, they found a \textit{moving backward} skill without access to extrinsic reward in the Half-Cheetah environment. \textbf{VIC} \cite{gregor2016variational} and \cite{binas2019journey} also did some similar experiments without exhibiting the same diversity of skills.

Three main limits of these approaches can be identified. Firstly, it is hard to see how these methods could be applied to environments that are different from MuJoCo, which seems adequate to these methods since the agent often falls in the same state (on the floor) where the state is hardly distinguishable. Secondly, the agent is unable to learn to generate goals without unlearning its skills. This way, the goal distribution generated by the agent has to stay uniform \cite{gregor2016variational,eysenbach2018diversity}. Thirdly, none of these approaches tries to use a continuous embedding of trajectories, such embedding could facilitate interpolation between skills, thereby would give more control to the inter-option policy.

\paragraph{Learning a continuous embedding:}
\textbf{DISCERN} \cite{warde2018unsupervised} tackles the last issue and considers the goal space as a state space. Then it does an approximation of $\log q(g|s_{final})$ by trying to classify the final state of the trajectory as the right goal among other goals selected from the same distribution as the real one. Intuitively, the agent learns to find the closest goal to the final state from a set of goals. The second limits mentioned earlier remain, they did not integrate their algorithm in a hierarchical structure and they implemented an \textit{ad-hoc} mechanism to maintain an uniform goal distribution in their buffer. An other way to use a continuous goal space with this objective is to use \textbf{DADS} \cite{sharma2019dynamics}. In DADS, the authors rather compute the reward as $\log q(s_{t+1}|g,s_t) - \mathbb{E}_{g\sim p(g)}\log q(s_{t+1}|g,s_t)$ which is also derived from mutual information between goals and states, but using the symetric opposite of entropy (Equation \eqref{information2}). The first term ensures goals are distinct, and the second term makes exploration easier. They also use model predictive control \cite{garcia1989model} (MPC) to plan on the behavior level. Their results show a considerable improvement of the quality of learned skills with respect to DIAYN. However, they rely on a stochastic parameterized distribution on the state space as predictive model and, as a result, it is not clear how well they could perform without the access to the $(x,y)$ coordinates on harder environment. They did not train their agent end-to-end.
An other way to provide a continuous embedding can be found in \textbf{SeCTAR} \cite{co2018self}; the authors propose to encode trajectories into a latent space, and to decode in the same way as an auto-encoder. The trajectories generated by the latent-conditioned policy and those of the decoder learn to be consistent with each other. To do so, the policy uses the likelihood of the trajectory (computed by the decoder) as a reward while the decoder learns like a LSTM-based VAE. As a result, the agent policy matches the distribution over trajectories learned by the decoder. The advantage of this approach is that it can take advantage of the decoder to use it as a forward model at the option level. Doing so, it manages to get interesting results on simple environments using a planning method. The major limitation is the use of recurrent neural networks, which are known to be computationally expensive, for both encoder and decoder and the fact that the learning process is not carried out end-to-end. Indeed, they first learn skills and only then use them in a hierarchical setting. Their decoder is particularly computationally ineffective for planning since it predicts the entire trajectory of the closed-loop option. 

\paragraph{Off-policy adaptation:}\citename{hausman2018learning} propose a way to learn theses policies with an extension of the Retrace algorithm \cite{munos2016safe} for off-policy learning, in the setting of multi-task learning. They manage to learn several trajectories for one task by learning a distribution of latent variable $p(g|t)$ where $t$ is the task. Although, after training on several tasks, learning a new distribution $p(g|t')$ can be enough to solve a new task $t'$, that is not a bottom-up approach; this is not studied in a hierarchical framework and pre-training tasks have to be related to the new task to find an optimal policy.

%It should be mentioned that some articles try to maximize a similar diversity goal with a predictive model and distance function \cite{song2018diversity}. However this is throwing away some advantages of this approach, which consists in avoiding a distance function and handling stochastic skills.

%se réduit au fur et à mesure que l'agent apprend à distinguer les \textit{options}.20

\begin{table*}
\centering

\begin{tabular}{|l|}
    %\hline
    %Methods which strive to find a  \\
    %\hline
    \hline
    \textbf{Goal space from the state space}  \\ 
    \hline
    \hline
    HAC \cite{levy2018hierarchical} \\    
    HIRO \cite{nachum2019data}\\
    FuN \cite{vezhnevets2017feudal}\\
    RIG \cite{nair2018visual}\\
    Skew-fit \cite{pong2019skew} \\
    SFA-GWR-HRL \cite{zhou2019vision}\\
    Sub-optimal representation learning \cite{nachum2019near} \\
    DISCERN \cite{warde2018unsupervised} \\
    \hline
    \hline
    \textbf{Mutual information between goals and trajectories}   \\
    \hline
    \hline    
    SNN4HRL \cite{florensa2017stochastic} \\
    VALOR \cite{achiam2018variational} \\
    DIAYN \cite{eysenbach2018diversity} \\
    VIC \cite{gregor2016variational} \\
    DISCERN \cite{warde2018unsupervised} \\
    SeCTAR \cite{co2018self} \\
    DADS \cite{sharma2019dynamics} \\
    Extension of Retrace \cite{hausman2018learning} \\
    %DEHRL \cite{song2018diversity} \\
    \hline
\end{tabular}
 \caption{Classification of methods which focus on learning a skill representation.}\label{tab:skill_abstraction}

\end{table*}

\subsubsection{Conclusion}

To summarize, there are two main groups of work about self-generation of goals. The first ensemble considers its objectives as states, the advantage is then to have a continuous space enabling interpolation. The disadvantage is that it requires to find the right comparison metric and the right way to compress the state space. Otherwise, the agent is not able to let the inter-option policy produces high-dimensional actions and is unable to discern similar states from different states. The second ensemble takes advantage of information theory to partition trajectories. The option space has a limited size but intra-option policies suffer from catastrophic forgetting, skills are more stochastic and interpolation between skills is harder. In addition, as highlighted by \citename{sharma2019dynamics}, using a well-built state space could also make skills more meaningful. Table \ref{tab:skill_abstraction} summarizes the classification of methods which learn a goal embedding.

\subsection{Curriculum learning}\label{sec:curriculum}

%These approachs have the benefit of avoid to hand-define some pre-training task as in \cite{heess2016learning}.

The aim is here to learn to choose an objective which is neither too hard nor too easy to speed up the learning of several goals of an agent. Specifically, this kind of work tries to \textbf{learn an efficient curriculum among the goals of an agent}, the counterpart is that these works generally assume more prior knowledge. To be efficient, the curriculum must avoid forgetting what an agent has previously learned and avoid trying to learn both unlearnable tasks and fully learned tasks.

So far we have seen that learning options improves both exploration when learned in a bottom-up way, and the credit assignment; that the use of motivations may help to build state feature space with specific and helpful properties; and that IM may guide the agent towards novel states. However, these methods are not incompatible with each other. Goal-parameterized RL could benefit from both getting an interesting representation of the state space and exploring at the inter-option policy level. Here, we emphasize some works at the intersection between these IMs. Particularly, the point is \textbf{how IM facilitates the exploration of a parameterized goal}. In this section, the goal space may be hand-engineered or learned.

In this section, goals are often derived from the state space and learned using Hindsight Experience replay (HER) \cite{andrychowicz2017hindsight}. We refer to Section \ref{uvfa} for an explanation on how we can learn goals while acting on other goals. We define 1- the choice of goals on which to learn as made possible by HER; 2- the selection of goals on which to act as a strategy to provide goals to the intra-option policy. This section is divided into four parts. The first part is dedicated to explain why the combination of skill abstraction and\textit{curriculum learning}improves the state space exploration. Then, we will analyze different methods for goal selection; 1-simple sampling methods; 2-modelling of the problem as a multi-armed bandit learning with learning progress; 3- application of adversarial techniques to produce state-goals.

\subsubsection{Curriculum learning on state related goal space for exploration}\label{curriculum_exploration}

%In this part, we study the interest of  %how and why both skill abstraction and\textit{curriculum learning}improve exploration. 
We can first notice that it is desirable to have a goal space strongly related to the state space, else, skills would be useless. In Section \ref{gen_goal}, the two major methods both build a goal space which has the property of mapping states to goals. This is salient when the goal space is directly derived from the state space (see \S\ref{sec:goal_state}); this is also a direct consequence of the definition of methods based on maximizing the mutual information between states and goals (see \S\ref{miskill}). It therefore happens that exploring the goal space is the same as exploring the state space. This is notably highlighted by experiments of \cite{sharma2019dynamics,co2018self,lee2019efficient}.

Let us take, as an example, an uniform sampling among a goal space which is a compressed version of the state space. For each sample, the agent will act in order to reach this state. This is due to the fact that goals can act as a latent variable defining the trajectory of the agent \cite{lee2019efficient}. When the sampling is uniform, the agent will try to go into all states. In fact, uniform sampling is closed to methods based on novelty-based IMs (\S\ref{sec:novelty} and \S\ref{sec:discrepancy}) since, in both cases, it approximately encourages the agent to generate, through its actions, an uniform state distribution. However, we do not mean to attenuate the role of curriculum learning; this is still required to drastically speed-up the exploration process. 

We will now detail methods for both sampling goals on which to learn and goals on which to act.

\subsubsection{Simple goal sampling}

Until now we have focused on IM as an intrinsic reward, however, this is not a general rule. For example, one can think of some simple strategies to choose tasks as long as the choice does not depend on an extrinsic reward. In this subsection, we study how such simple strategies can be efficient.

 \citename{andrychowicz2017hindsight} fully take advantage of \textbf{HER} and experimented different ways to sample goals on which to learn from trajectories. First the agent randomly sample a transition, then it replaces the initial goal by another one. They propose four strategies to choose the replacing goal:
\begin{itemize}
\item %The final state of the same episode as the transition being replayed.
The final state of the episode whose transition is being replayed.
\item Random goals originating from the episode whose transition is being replayed.%the same episode as the transition being replayed.
\item Sampling randomly from the buffer.
\item States arising after the transition being replayed in the same episode.
\end{itemize}

It appears that using future states or final states are the best working goals and generalization over the goal space pushes the agent towards its main goal. This is probably because these states act as a novelty bonus, helping the policy to generalize over the goal space and learn beyond its current abilities. In fact, count-based methods from Section \ref{sec:novelty} also reward the agent when it reaches a state it never went into: both methods have similar side-effects. The advantage of sampling methods compared to other contributions (\S\ref{sec:multi-armed} and \S\ref{sec:adversarial}) is that the agent continues to try to reach its true-goal state while performing exploration in the goal space. Few works extended HER while remaining in the field of IM. \textbf{Prioritized HER} \cite{zhao2019curiosity} proposes to adapt prioritized experience replay \cite{schaul2015prioritized} by over-weighting rare states. We can see this idea as an heuristic to consolidate novelty-based sampling. They slightly improve the results over HER at the cost of maintaining a density model. 

Even though these methods learn with new sampled goals, they act based on an extrinsic goal to solve. Therefore, they require a goal parameterized target. To improve exploration without an extrinsic parameterized goal, \textbf{UNICORN} \cite{mankowitz2018unicorn} samples uniformly in a goal space to interact with. This strategy can be effective since new goals and the generalization ability of the agent can make it go toward boundaries of its skills. However, it is unclear how the agent could behave in a poorly constructed goal space (such as a pixel state space).

\subsubsection{Task selection as a multi-armed bandit}\label{sec:multi-armed}

A common way to choose the task is to associate a learning progress value to each task. The learning progress value for a task evaluates the improvement of the agent in doing this task. %Learning progress rewards the agent only if it is making progress. L
Learning progress has several attractive properties:
\begin{enumerate}
    \item If the agent master a task, it does not improve itself and stops learning the task.
    \item If the task is unlearnable, the agent does not persist to achieve it.
    \item The agent focus on the task which mostly matches its level.
\end{enumerate}  

It is generally defined as the first order derivative of the performance:
\begin{equation}
    R_{int}(o_T)=\frac{\partial R_{o_T}}{\partial T}
\end{equation}

where $o_T$ is a task and $T$ is the number of times a task has been chosen. 

\textbf{Teacher-Student} \cite{MatiisenOCS17} models the problem of choosing a task as a non-stationary multi-armed bandit which aims to improve the learning progress of a task-specialized policy. The agent chooses a task among a set of pre-defined tasks using its estimated learning progress. Then it tries to solve the task with a task-specific reward. The authors propose to evaluate the learning progress with the coefficient of the slope of the reward progress computed with a simple linear regression through recent interactions. However, tasks are just a different setting of the same carefully designed objective. For example, the agent has to reach a similar target in an increasingly larger labyrinth rather than going towards a target to pick up an object to use it somewhere else. It makes the generalization of the policy easier.
\textbf{CURIOUS} \cite{colas2019curious} models the problem likewise with diverse hand-made tasks (e.g, grasping a cube, pushing it into an other). Each task has its goal space (e.g different configurations of cubes), making it possible to integrate HER; furthermore, goals on which to act are sampled uniformly among the goal space. The learning progress over tasks is computed as the difference of rewards earned between two evaluation step, where an evaluation step consists of the mean reward obtained from some  previous episodes of the task. Even if the agent manages to learn across different tasks, it can take advantage of HER as long as tasks rewards overlap, or are even close to each other. 
%\textbf{IMGEP} \cite{abs170802190} also describes a RL framework with a hand-made goal space where the agent selects the goal with a similar learning progress measure.
In fact, CURIOUS is part of a general framework \textbf{IMGEP} (Intrinsically Motivated Goal Exploration Processes) where the selection of goals is modeled as a Multi-armed bandit. Within this framework, the goal space can be hand-made \cite{abs170802190} or learned through a VAE  \cite{pere2018unsupervised,reinke2019intrinsically} as in Section \ref{gen_goal}. However, most works of this framework do not take advantages of DRL.
\textbf{CLIC} \cite{fournier2019clic} extends and improves over CURIOUS by no longer considering some predefined tasks but using an other intrinsic reward to manipulate objects. This reward is computed as the distance, for only one specific feature or object, between the current state and the final state of the agent. However, this work is based on the assumption that the state space is disentangled, i.e. each feature corresponds to the state of one object.

%\textbf{M-GRAIL} \cite{santucci2019autonomous} treats interrelated tasks by modelling the problem as Markovian and adding information on already solved goals into the high-level state. There are several issues: the agent learns a different policy per task, relies on low-level reward prediction and generalization to more diverse tasks is not clear.

Among all these works, it appears that computing the learning progress is difficult and requires an evaluation step. Although not used with an RL agent, \citename{graves2017automated} propose multiple other methods to compute the learning progress for learning to choose a task by leveraging the distribution model of a stochastic neural network. In particular, they introduce the variational complexity gain, which can be measured as the difference between two consecutive KL-divergence between a fixed prior and the posterior over the parameters of the model (the task learner). It is in fact very close to the information gain defined in \S\ref{sec:info_gain}. In the same way, \citename{linke2019adapting} introduce the weight change with an adaptive learning rate. A large scale study of the best way to choose the intrinsic reward for a multi-armed bandit agent trying to learn simple tasks can be found in \citename{linke2019adapting}.

These works associate a learning progress value to the task in order to choose the task. However this is harder when the task is continuously parameterized; it could require to partition the state space \cite{baranes2013active}. The next section show different ways to make the association, at the cost of several assumptions.

\subsubsection{Task selection with adversarial training}\label{sec:adversarial}

Here, we explain how adversarial training can overcome the need of a discrete goal space to use the learning progress. In the paradigm of adversarial training, two modules face each other: the first one, the generator, tries to fool the second, the discriminator, which must avoid to be mistaken. As they progress, the generator proposes more and more convincing data whereas the discriminator is getting harder and harder to fool. The works in this category use a goal space related to the state space; therefore, these works can be used to improve exploration.

\textbf{GoalGAN} \cite{florensa2018automatic} learns to generate increasingly complex continuous goals with a Generative Adversarial Networks (GAN) \cite{goodfellow2014generative} in order to make the policy progressively learn to go everywhere. The generator of the GAN learns to produce goals of intermediate difficulty and the discriminator learns to distinguish these goals from others. Intermediate difficulty is characterized as a goal that an agent achieves from time to time. In this article, the intra-option intrinsic reward relies on an hand-engineered indicator function which attributes a binary reward if the agent is close to a goal. The parameterized goal space is assumed to be known, for example, the article present the goal space as the $(x,y)$ coordinates whereas the state space is larger. Similar issue can be found in \textbf{Self-play HRL} \cite{sukhbaatar2018learning}. In this article, adversarial method is applied to learn a goal space which is a compressed state space. During a pre-training step, an agent (generator) tries to go into the state that an other agent (discriminator) went into, whereas the discriminator learns to go into areas the generator cannot reach. In other words, the generator tries to produce trajectories that the discriminator can not differentiate from its own. Thus, the reward of the generator is one if a distance function between the discriminator final state and its own final state is under a predefined threshold; the reward is reversed for the discriminator. Due to the used architecture, the generators implicitly learn to compress the goal state. This goal space can be used to solve an afterwards task in a hierarchical way. Although this method seems to manage harder environment than GoalGAN, it is also limited by the need of an expert for designing the reward function.

To avoid the assumptions of a simple goal space, \citename{venkattaramanujam2019self} try to capture environment dynamics in a distance function (see \S\ref{sec:statemeasure}). They propose goal states separated by random actions from already reached states. It seems to perform similarly to GoalGAN, but needs the ability to reset the agent in all states of the environment.

However, one major drawback of using adversarial methods is that the agent can not focus its exploration on areas useful to an external task.

\begin{table*}
\centering
%\begin{threeparttable}
\begin{tabular}{|l|}
    \hline
    \textbf{Goal sampling}\\ 
    \hline
    \hline
    HER \cite{andrychowicz2017hindsight} \\
    Prioritized HER \cite{zhao2019curiosity} \\
    UNICORN \cite{mankowitz2018unicorn} \\
    \hline
    \textbf{Multi-armed bandit} \\ 
    \hline
    \hline
    Teacher-Student \cite{MatiisenOCS17} \\
    CURIOUS[task sampling]\cite{colas2019curious} \\
    %IMGEP \cite{abs170802190} \\
    %M-GRAIL \cite{santucci2019autonomous} \\
    CLIC \cite{fournier2019clic} \\
    \hline
    \hline
    \textbf{Adversarial training} \\
    \hline
    \hline   
    GoalGAN \cite{florensa2018automatic} \\
    Random actions \cite{venkattaramanujam2019self} \\
    Self-play HRL \cite{sukhbaatar2018learning} \\
    \hline
\end{tabular}
  %\begin{tablenotes}
   % \item[1] End-to-end learning.
  %  \item[2] Goal space.
 % \end{tablenotes}
%\end{threeparttable}
 \caption{Classification of task selection methods for building a curriculum of skills.}\label{tab:curriculum}
%Ckassification of task selection methods for building a curriculum of skills
\end{table*}

\subsubsection{Conclusion}

To conclude, exploration can also be determinant in a goal space. We identified three methods for doing that summarized in Table \ref{tab:curriculum}. The first one uses simple goal sampling methods which present the advantage of being simple and learnable in an end-to-end way. The second one models the problem of choosing a task as a multi-armed bandit reinforced with learning progress. It allows to bypass poorly built goal space and speeds up learning, but is hardly computed with a continuous goal space. The last one uses adversarial training to generate adequate tasks. Adversarial methods can learn with a continuous goal space but need a good goal representation and cannot be learned in an end-to-end way. Globally, it has been shown that\textit{curriculum learning}methods could significantly accelerate the skill acquisition and allows to more efficiently explore the state space when skills are related to states. However most of this work relies on strong assumptions to measure the accomplishment of the option. We believe that further work will have to relax these assumptions. An interesting inspiration could be taken from Powerplay \cite{schmidhuber2013powerplay}. This is a theoretical and global framework beyond such assumptions which is continually searching for new tasks, however it still lacks a concrete application.

\section{Limitations and challenges of the methods}\label{limite}

Much work is limited by challenges out of the scope of RL, such as the performance of density models \cite{bellemare2016unifying,ostrovski2017count} or predictive models \cite{nachum2019data,nair2018visual}, or the difficulty to approximate mutual information between two continuous random variables \cite{gregor2016variational}. These limitations are beyond the scope of this article and we focus here only on challenges related to RL. Despite the heterogeneity of work on IM in RL and the specific limitations of each method, we select and present in this section five major issues and challenges related to these approaches.

\subsection{Environment stochasticity}\label{sec:stochasticity}
%\begin{enumerate}

%We saw in the previous part that it was interesting to maximize the compression progress and that most of the work were relative to information compression, and not to compression progress. 
A lot of works in Section \ref{curiosity} (related to exploration problem) create their reward with \textit{prediction error} instead of the improvement of prediction error (see \S\ref{sec:compression} for a thorough analysis). This discrepancy explains the difficulty of several works to handle the white-noise effect \cite{burda2019largescale}  or, more generally, to handle the stochasticity of the environment. 

Some articles from the state of the art handle this issue (see Table \ref{tab:curiosity}), but each of them has drawbacks:
\begin{description}
\item[ICM] \cite{pathak2017curiosity} can not differentiate local stochasticity from long-term control.
\item[Count-based methods] \cite{ostrovski2017count,bellemare2016unifying,tang2017exploration,martin2017count,burda2018exploration} can only handle one kind of stochasticity. For instance, let us assume that one (state, action) tuple can lead to two very different states with 50\% chance each. The algorithm will manage to count for both states the number of visits, although it would take twice as long to avoid to be too much attracted. However, if the next state is a new randomly generated one every time, it is not clear how these methods could resist the temptation of going into this area since the counting associated to this state will never increase. %We characterize such fully random next states as non-regular stochasticity, as opposite to the regular one.
\item[State comparison] \cite{savinov2018episodic,fu2017ex2} relies on a large number of comparison between states. It is unclear how it could scale to a larger number of states. \citename{kim2019curiosity} can avoid distractors at the condition that the agent finds extrinsic rewards.
\item[Information gain] \cite{houthooft2016vime,achiam2017surprise,shyam2018model} seems particularly adequate to handle stochasticity since the agent only considers the reduction of uncertainty about dynamics, and this reduction is related to the degree of stochasticity. However it is difficult to apply in practice as evidenced by its score or its additional computational modules (see Table \ref{tab:curiosity}). %This difficulty comes from the fact that an update is responsible for only a small change in parameters and is dependent of updates in other states. It results that intrinsic rewards may have a high variance. 
Similarly, related exploration methods (c.f \S\ref{sec:multi-armed}) in the goal space use a similar motivation, denoted as learning progress. Although this is efficient to avoid stochasticity, it only handles a discrete goal space.
\item[Goal sampling] \cite{andrychowicz2017hindsight} has not, to our knowledge, been confronted yet to noise in the state space. It would probably be resilient. However, it requires a parameterized extrinsic goal to guide exploration.
\item[Adversarial goal selection] is sensitive to stochasticity since the discriminator would maximize its objective when sampling purely \cite{sukhbaatar2018learning} or semi \cite{florensa2018automatic} stochastic goal areas.
% We saw that it was particularly efficient. We comes back to this point in Section \ref{sec:exploration_curriculum}.

\end{description}

Additionally, \citename{burda2019largescale} highlight that, even if an environment is not truly random, the agent can get stuck in some parts of the environment. To illustrate this, the authors placed a television in their 3D environment and added a specific action to randomly change the picture of the displayed picture. It turns out that their agent (ICM and prediction with random features) kept looking at the picture. It would be interesting to test a broader class of algorithms in order to test their abilities to handle such a difficult stochastic setting. In fact, there is a lack of distinction between stochasticity in the environment and uncertainty relative to environment dynamics, although the agent must act differently according to these two types (we will discuss it again in Section \ref{sec:free_energy}). Information gain methods currently present the most serious outlooks to solve this issue. 

\subsection{Binding skills learning and exploration}\label{sec:binding}

For two reasons, we claim that skill learning can be an important source of improvement for the exploration issue. We already have investigated the direct interest of skill learning to explore (\S \ref{sec:abstraction}) and found that it can reduce the noise of standard exploration methods resulting in a faster access to sparse rewards. This aspect is further developed in the next subsection. In addition to that, we saw that skill learning makes the credit assignment more effective and faster. This is extremely important since an intrinsic reward can be a fast moving non stationary reward. If the long-term attenuation parameter $\gamma$ is high, such a reward function could propagate along different states very slowly since the state sequence between the state we want to value and the rewarded state is very long. It results that the policy is improved very slowly.
This is why pseudo-count methods use a mixed Monte Carlo update \cite{ostrovski2017count}, which consists in using a soft interpolation between Monte Carlo and Bellman equation (Equation \ref{eq:bellman}) to update values. However their method only partially solves the problem on the cost of a higher variance.

In a different way, if the fast moving non stationary intrinsic reward changes an abstract policy option, it can propagate to every states much faster without any additional cost. To illustrate this, let us assume that options of length 20 are available, and that the target state (with the highest intrinsic reward) is 1000 states away from the initial state. In a tabular setting, it would take at least 1000 updates with a $\gamma$ of 0.998, whereas it would take only, at least, 50 updates to the option policy with a $\gamma$ of 0.98. 
 
\subsection{Long-term exploration}

\begin{figure*}
\begin{center}
\includegraphics[width=10cm]{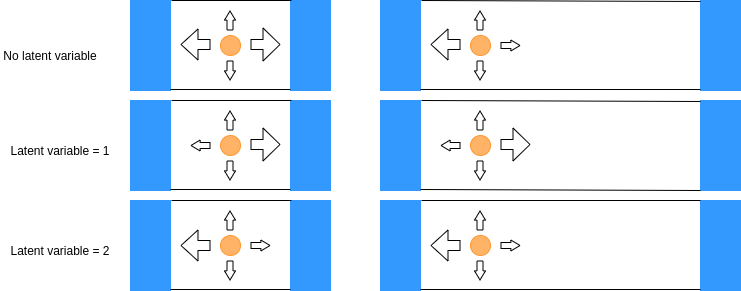}
\caption{Illustration of the benefit of using latent variable for exploration. The orange circle wants to explore its two environment through cardinal movements and the blue area characterize unexplored spaces. At the top, the agent uses standard novelty-based exploration strategies. In the middle, the agent explores conditioned on a latent variable initialized to one. At the bottom, the agent explores conditioned on a latent variable initialized to two. The latent variable forces the agent to separate the area it goes into.}
\label{im:detachment}
\end{center}
\end{figure*}
 
To our knowledge, none of the existing approaches handles long-term information search.
The most challenging used benchmarks in the current state of the art are \textit{DMLab} and \textit{Montezuma's revenge}, yet very sparse reward games such as \textit{Pitfall!} are not currently addressed and should be investigated. In \textit{Pitfall!}, the first reward is reached only after multiple rooms where it requires specific action sequences to go through each room. State of the art on IM methods \cite{ostrovski2017count} achieve 0 mean reward in this game. At the opposite, imitation RL methods \cite{aytar2018playing,hester2018deep} are insensitive to such a specific reward, and thus, exceed IM methods with a mean reward of 37232 on \textit{Montezuma's revenge} and 54912 on \textit{Pitfall!}. Even though these methods use expert knowledge, this performance gap exhibits their resilience to long-term rewards. Compared with intrinsic reward methods, which do not exceed a 10000 score on \textit{Montezuma's revenge} \cite{burda2018exploration} and hardly achieve a score on \textit{Pitfall!} \cite{ostrovski2017count}, it shows that IM is still far from solving the overall problem of exploration.

Furthermore, we want to emphasize that the challenge is harder when the intrinsic reward itself is sparse \cite{burda2018exploration}. In \textit{Montezuma's revenge}, it is about avoiding to use a key too quickly in order to be able to use it later. In every day life, it can be about avoiding to spend money too quickly. In fact, it looks like there is an exploration issue in the intrinsic reward function. Intrinsic reward can guide the exploration at the condition that the agent finds this intrinsic reward. There may be two reasons causing the intrinsic reward to be sparse. The first is partial observability, with which most models are incompatible (see also Section \ref{sec:first_view}).
The second, as identified by \citename{goexplore}, is the result of a distant intrinsic reward coupled with catastrophic forgetting and action stochasticity. In fact, it may occur that the policy hesitates between different modes of intrinsic rewards, resulting in either a noisy exploration policy or a forgetting of a mode. While this challenge could be solved with an approach using planning methods \cite{hafner2018learning}, we want to emphasize the potential of goal-parameterized RL. The agent could explore through the goal space as in Section \ref{sec:curriculum}, enjoying its benefits. Using latent variables enables to capture different novelty modes, stabilizing exploration. This is illustrated in Figure \ref{im:detachment} where the orange circle strives to explore with two intrinsic rewards modes. When using standard-novelty based methods (at the top), the agent either has a noisy policy which can potentially making it go back and forth (left) or it forgets about one mode and focuses only on a part of the environment (right). As shown at the bottom of Figure \ref{im:detachment}, using a latent variable forcing it to go in different part of the environment avoids the issue; in the middle, it can focus on exploration to the right while at the bottom, it concentrates on exploration on the left.

\subsection{Building a practical state representation}\label{sec:staterepr}

There are several properties that a state representation should verify. As humans, we are aware of distance between states, we can easily segment objects, perceive their position and abstract them, understand objects affordance (i.e. potential high-level actions made possible by the properties of the item) \cite{thill2013theories}. We are also aware of our spatial position in the world on several spatial scales. In addition, to make decisions, we easily use hidden state information such as past actions or past observations. Our state representation is rich, and enables us to get goal-directed behaviors or object-directed exploration. Such abstractions is the foundation of our cognition, but they are still missing in IM approaches. This limitation is particularly salient throughout our survey. We have already seen that building a good feature space is important for discovering goals, in order to compose with a reduced goal space \cite{nair2018visual} or to get object-oriented behaviours \cite{kulkarni2016hierarchical}. It is particularly highlighted in the work of \citename{eysenbach2018diversity} and \citename{sharma2019dynamics} where an access to the $(x,y)$ coordinate strongly improves the quality of behaviors. It is also crucial in works related to knowledge acquisition to get a significant \textit{prediction error}. For example, ICM \cite{pathak2017curiosity} proposes an interesting state representation restricted on what can be immediately controlled by the agent; EMI \cite{pmlr-v97-kim19a} manages to construct an embedding space where a simple linear forward model is adequate but without a specific structure; Sub-optimal representation learning \cite{nachum2019near} finds out coordinates of the agent \dots

There is a large literature on learning representations \cite{lesort2018state}, yet, there is currently few work which benefits from the recent advances in this area. While, on the other side, some work takes advantage of IM to learn representation spaces (see Section \ref{sec:staterep}), we strongly believe that option policies or exploratory policies can take advantage of such representation spaces; For example, \cite{kulkarni2016hierarchical} takes advantage of a predefined object-centered representation to achieve good scores on benchmark such as \textit{Montezuma's revenge}. As an other example, \citename{thomas2017independently} tries to learn a disentangled state space, whereas this is a prior knowledge in CLIC \cite{fournier2019clic}. The existing works could be cross-fertilized to overcome their current limitations.

It results that state representation and IM may be more and more intertwined and raise new questions: what mechanisms underpin the relationship between these two concepts ? Given that it is a chicken-and-egg problem, which one is learned first ?
 
\subsection{Decorrelating the goal learning process from the task}\label{sec:lifelong}
 
The advantage of decorrelating the learning of objectives from the learning of the task is to favor exploration and transfer learning. This is usually called \textbf{bottom-up learning} because skills are learned before the task. Typically, it can be positive for an agent to learn to walk before learning to reach an object (which is the extrinsic task); then it can reuse this walking behavior to fulfill other tasks. As a result, \textit{bottom-up} learning can improve both exploration and \textit{transfer learning}. If this learning process has made significant progress, it is still difficult to learn simultaneously tasks and skills without enduring catastrophic forgetting \cite{mccloskey1989catastrophic,florensa2018automatic}. Indeed, when the agent sequentially learns tasks, it forgets the first task while learning the next one. Some work already tackles the catastrophic forgetting problem  \cite{kirkpatrick2017overcoming,parisi2019continual} but it has not, to our knowledge, been evaluated with IM and on a large number of tasks. For example, \citename{colas2019curious} showed that learning progress based\textit{curriculum learning}could alleviate the issue on a small number of tasks. However, even if one would generalize the learning progress to continuous goal space, this approach may not scale to a large number of goals. The risk is that the agent concentrates its resources on avoiding catastrophic forgetting instead of learning new tasks. More broadly, these aspects are part of continual learning \cite{parisi2019continual}, i.e. the agent's ability to continually train and keep improving skills throughout its lifespan.
%continually acquire, fine-tune,
%and transfer knowledge and skills throughout their lifespan
\section{Review of tasks involving IM}\label{tasks}

We identified four fundamentally different types of tasks on which IM methods are tested. In this subsection we emphasize their particularities and the solving algorithm proposed in the literature.

\subsection{Locomotion}

Locomotion tasks are mostly related to MuJoCo environments such as \textit{ant} or \textit{humanoid} where the goal of the task is to move an agent \cite{duan2016benchmarking}. Most related work consider exploration and skill acquisition methods. Exploration methods only solve easy locomotion tasks, e.g. Half-Cheetah having a 17-dim observation space and 6-dim action space \cite{houthooft2016vime,pmlr-v97-kim19a,fu2017ex2}. On the other side, skill acquisition methods manage to learn to move forward (by crawling or walking) on harder morphologies, e.g. \textit{Ant} having a 111-dim observation space and a 8-dim action space \cite{achiam2018variational,eysenbach2018diversity}. Interestingly, a diversity heuristic without extrinsic reward suffices to get representations of different interesting skills. It suggests that diversity heuristic could be enough to handle proprioceptive incoming data. However, currently, too much useless skills are learnt and they can not be used while being learnt.

\subsection{Manipulation}\label{sec:manipulation}

Manipulation tasks can be about moving, pushing, reaching objects for a movable robotic arm. Few exploration methods have been tested \cite{lee2019efficient,pathak2019self} and they only manage to touch and move some objects. It is particularly interesting for skill acquisition methods \cite{hausman2018learning,nair2018visual} but this is not actually a major focus since it lacks object-oriented objective (as argued in \S\ref{sec:staterepr}). It is a standard task for\textit{curriculum learning}algorithms \cite{colas2019curious,santucci2019autonomous} since, for example, an agent has to learn to reach an item before moving it.\textit{curriculum learning}algorithms can be very efficient but at the cost of a hand-made goal space.

\subsection{Navigation}\label{sec:navigation}

Navigation tasks are about moving an agent in a maze. This is the broadly tested task and includes every kind of methods we presented. It can consist in moving a MuJoCo \textit{ant} or \textit{swimmer} in order to pick up food or to reach a target area. In the same way, Atari games generally consist in moving an agent into a rich environment, but with simpler discrete action space. Similarly to manipulation tasks, it requires target-oriented behaviors and favors the use of skills as states rather than diversity heuristic (despite a lot of progress in this way made by \citename{sharma2019dynamics}). Exploration methods are particularly efficient in discovering new areas and make sense, but are brute force and could be considerably improved as discussed in Sections \ref{sec:binding} and \ref{sec:staterepr}. Results of exploration through curriculum (\S\ref{curriculum_exploration}) also showed to be a nice alternative to standard exploration methods (\S\ref{sec:adversarial}) because of\textit{curriculum learning}capacity to capture different reward mode (\S\ref{im:detachment}).

\subsection{First-person view navigation}\label{sec:first_view}
First-person view navigation tasks are particularly challenging since the agent only receives a partial first-person visual view of its state and must learn its true state (e.g. its position). There are few work addressing these environments, mostly for exploration \cite{pathak2017curiosity,savinov2018episodic,fu2017ex2}, but they manage to efficiently explore the environment \cite{savinov2018episodic}. There is a lack of an application of count-based methods showing whether partial observability is a drag for the method. To the best of our knowledge, there is no work that tackle these environments in skill learning methods. It suggests a large need for a low-cost way to build the true state of the agent from partial observations. Yet, it is also not tackled in state representation learning methods.

Nevertheless, standard RL methods could take advantage of breaking down the partial observability into a long-term one at the higher level of the hierarchy, and into a short-term one at a lower level of the hierarchy. It could make the training of a recurrent neural network easier by reducing the gap between a notable event and the moment one needs to retrieve it in memory to get a reward. For example, in a 3D maze where the agent tries to reach an exit, a long-term memory could memorize large areas the agent went into whereas the short-term memory could focus on short time coherent behaviors.

%\subsection{Conclusion}

%It appear through our study that intrinsic motivations are different roles on tasks.

\section{Analysis}\label{analyse}

In this section, we highlight the global direction taken by works on intrinsic reward. We first stress out the role of mutual information, which leads us to briefly review the idea of  information compression and the way it relates to works we reviewed. Then, we exhibit prior knowledge induced by current compression algorithms. It suggests that \textit{a priori} different approaches are similarly composed of a compression information module rewarding a reinforcement learning algorithm; in fact, each classes of our classification (See Table \ref{tab:tableofcontents}) is a specialized component of this unified view. We further emphasize this aspect by noting the apparent incompatibility between the \textit{free-energy} principle and intrinsically rewarded agents. We terminate by showing that the combination of these components could form a complete developmental architecture. This architecture may be able to solve in an end-to-end way all the tasks studied in Section \ref{tasks}.

\subsection{Mutual information as a common tool}

A redundancy seems to appear throughout the whole study, whether it is on knowledge acquisition or skills learning. Mutual information seems to be central to expand abilities of the agent, so we briefly review existing relations between mutual information and intrinsic rewards.

\subsubsection{Direct use of mutual information}

We have first seen that \textit{empowerment} is entirely defined with mutual information  (cf. \S \ref{empowerment}). Similarly, a whole section of work in \S \ref{miskill} is based on mutual information between the path resulting from the goal and the goal itself. VIME \cite{houthooft2016vime}, AKL \cite{achiam2017surprise} and MAX \cite{shyam2018model} maximize information gain, i.e. the information contained in the next state about the environment model $I(s_{t+1};\Theta|,a_t)$ where $\Theta$ are the parameters of the forward model. At last, EMI and CB \cite{pmlr-v97-kim19a,kim2019curiosity} make use of mutual information to compute the state representation. Although it is not an IM work, \citename{still2012information} suggest that the agent has to maximize mutual information between its action and the next states to improve its exploration policy.

\subsubsection{Function equivalent to mutual information}

\textit{Prediction error} \cite{nachum2019near,nachum2019data,pathak2017curiosity} is also related to mutual information \cite{de2018unified}, since it is very close to information gain methods. In the same way, they try to maximize the information that a forward model contains about its environment but are limited by its inability to encode stochasticity.

In addition to that, \citename{nachum2019near} explain that their method learns a state representation maximizing mutual information between the state in question and the next states. At last, \citename{bellemare2016unifying} show that rewards which come from \textit{pseudo-count} \cite{bellemare2016unifying,ostrovski2017count} are related to the one from information gain.

Finally, as noted by \citename{alemi2016deep}, the VAE objective is a specific case of the variational information bottleneck, which fully relies on two mutual information terms. It results that most work using this type of autoencoder is using the same mutual information tool (e.g \cite{klissarovvariational,co2018self}). 
%In fact, their agent acts to put the real state density into their model, which is impossible since the agent goes more often through state closed to his initial state. But still, it looks like they want to maximize mutual information between next state and the density $I(s_{t+1};\Theta)$ where $\Theta$ are parameters of the density model.
 
%thomas2018disentangling utilise aussi l'information mutuelle

\subsection{IM as information compression}\label{sec:compression}

\citename{schmidhuber2008driven} postulates that the organism is guided by the desire to compress information it receives. Therefore, the more we manage to compress received data from the environment, the higher the intrinsic reward is. Nevertheless, he indicates that this is the improvement which is important, and not the compression degree in itself, or an agent could decide to stay inactive in front of noise or an uniform darkness. As noticed by \citename{schmidhuber2007simple}, a breakthrough in compression progress is called a \textit{discovery}.

Data compression is strongly related to the observation of regularities in these very same data. For example, what we call a face is, in our environment, an ensemble appearing in a recurrent basis and composed of an oval shape containing two eyes, a nose and a mouth. Likewise, a state of the environment can be described with only some of the most pertinent features. Emphasizing this aspect makes this paradigm close to the minimum description length principle \cite{grunwald2007minimum} which considers learning as finding the shortest description of data. In our case, it implies that IM results in a search for new regularities in the environment. It is particularly salient in the works we reviewed.

It has been shown that methods on information gain are directly linked to information compression progress \cite{schmidhuber2008driven,houthooft2016vime}. ECO \cite{savinov2018episodic} tries to encode the environment by storing as more diverse states as possible; CB \cite{kim2019curiosity} and VSIMR \cite{klissarovvariational} direct towards the least compressed states. Predictive models \cite{burda2019largescale} encode environment dynamics in a parameterized model (often a neural network). The \textit{empowerment} is similar, it should be recalled that this is about directing an agent towards areas in which it has control, i.e. in which states are determined by agent actions. It is possible to reformulate \textit{empowerment} as the interest of an agent for areas where its actions are a compression of the next states. Indeed, \textit{empowerment} is maximal if every path leads to their own states (always the same in the same order) distinct from those of other trajectories whereas it is minimal if all trajectories lead to the same state. Some work on skill abstraction explicitly tries to compress trajectories into a goal space. If they use the state space as goal space, we saw that the challenge was to correctly compress the space into a usable one. This leads to a part of work which rely on the quality of compression of the state space \cite{vezhnevets2017feudal,nachum2019near,pathak2017curiosity}. Another part of works rather focus on whether trajectories are distilled in a policy, allowing to compute learning progress \cite{florensa2018automatic,colas2019curious,MatiisenOCS17}

This enhances the fact that current lines of works in IM are about rightly choosing the data to compress and the way to compress it. 

\subsection{Prior knowledge}\label{sec:prior}

To summarize, investigated models often have as common point to be composed of two modules: 1- The first module is a policy maximizing the intrinsic reward coming from the second module; 2- The second module computes the intrinsic rewards with a compression function, often related to mutual information between two variables.

This \textbf{compression function} is often implemented with neural networks in order to generalize across large state space. By finding causalities, the agent manages to compress the data it receives.

This study on the causality between data is possible because the works use \textbf{prior knowledge} on the structure of the data, i.e. the structure of the world. Used functions assume that data can be compressed. We have identified several types of prior knowledge:
\begin{itemize}
    \item the environment is not entirely stochastic;
    \item the environment is fully deterministic (see \S \ref{sec:stochasticity});
    \item the environment is markovian with a large number of different states (trajectories compression);
    \item an observation is composed of several independent features (state representation);
    \item actions can act as a metric in the state space (state representation);
    \item there is a hierarchical structure inside available tasks or accessible states (curriculum learning).
\end{itemize}

In fact, there is here a strong analogy with works on state representation \cite{lesort2018state}, which often incorporates reasonable assumptions to build a usable representation (e.g \citename{jonschkowski2015learning}, \citename{jonschkowski2017pves}). Using this knowledge is not necessarily negative since it is about the structure of the world.

To briefly summarize, a top view shows that one module takes advantage of a very global and task-independent knowledge on the structure of the world, through tools such as information theory, to compress incoming data. A measure of this compression serves as an intrinsic reward to enhance the reinforcement algorithm.

%Est-ce que cette sous partie est vraiment utile ?
\subsection{Free-energy principle}\label{sec:free_energy}

Unlike previous methods, the free-energy principle \cite{friston2010free,friston2009free,clark2013whatever} estimates that a common principle governs the learning of a predictive model and the choice of actions: the agent tries to reduce its surprise. This way, the actions have to be chosen so as to avoid any prediction error. Typically, it can explain some social behaviors in the infant such as imitative behaviors \cite{nagai2019predictive,triesch2013imitation}. Similar idea is exploited through the name of active efficient coding \cite{zhao2012unified,zhu2017joint}: the agent acts in order to get compressed sensory experiences. This framework has been proved to be effective to model eye movements such as vergence eye movements and stereo disparity perception \cite{zhao2012unified} or smooth pursuit eye behaviors \cite{vikram2014autonomous}.

A side effect is that an agent staying motionless in the darkness would minimize its prediction error. \textit{A priori}, it is difficult to determine how this principle could be compatible with actual methods. In fact, what lies behind most works we studied in Section \ref{curiosity} on exploration is an adversarial perspective: a module learns to decrease an evaluation function while a reinforcement algorithm pushes the agent towards difficult areas challenging the first module. 

It is more ambivalent in \S\ref{miskill} on skills learning. The goals are learnt by maximizing the information conveyed by the trajectory on the goal, but it is precisely the fact of choosing uniformly goals against this principle which allows learning (see Section \ref{miskill}). On the other side, the learning module and reinforced agent maximize the same objective, which is the probability of being in the chosen option knowing states covered by the intra-option policy.

\citename{schwartenbeck2019computational} differentiate two types of ambiguity: the first one is the uncertainty about the hidden state of the environment, the second one is the uncertainty about the model parameters. In other words, an agent can be certain about the uncertainty of the world. An agent should try to disambiguate this hidden state via \textit{active inference}, i.e. find observations where accounting for what to do. For example, an agent should check if there is a hole in front of him by knowing if there is a chance there might be one. On the opposite, active learning pushes the agent towards regions where the agent can gain information about the world, for example, the agent will be incited to push a button if it does not know what this button does. To our knowledge, reinforcement learning has still not been directly applied to such setting.

\subsection{Towards developmental learning}

We have seen in Sections \ref{motiv} and \ref{skill_learning} how IM enables to overcome multiple issues. Until now, we focused on single learning problems with respect to DRL. However, one would like a more general guideline making our agent more intelligent and efficient to solve the tasks presented to it. This way, several works address issues reviewed in this paper through a more systemic fashion \cite{santucci2016grail,pere2018unsupervised}. As noticed in \citename{guerin2011learning}, a safe path to build intelligence is to follow human development, that is what we call a \textbf{developmental approach}. A developmental architecture is based on the agent's embodiment which postulates that an agent must be grounded in its environment through sensorimotor capacities \cite{ziemke2016body,brooks1991intelligence}. The model we described in \S\ref{sec:modelRL} is in line with this principle. According to \citename{brooks1991intelligence}, \textit{everything is grounded in primitive sensor motor patterns of activation}. This \textit{everything} refers to the structure of the world and agent affordance; this is exactly what our first module (\S\ref{sec:prior}) strives to find out by compressing data it receives (\S\ref{sec:compression}). In fact, we can notice that all challenges of DRL tackled by IM are the one addressed by developmental learning.

More precisely, developmental learning refers to the ability of an agent to develop itself in an open-ended manner \cite{oudeyer2007intrinsic}; it is especially related to the autonomous learning of increasingly more complex and abstract skills by interacting with the environment \cite{guerin2011learning}. There are several key components in such a developmental process: the agent has to form concept representations and abstract reusable skills \cite{weng2001autonomous}, use it as a basis for new skills \cite{prince2005ongoing}, explore the environment to find new interesting skills, autonomously self-generate goals in accordance with the level and morphology of the agent. All these key components of a developmental process find a match with the RL issues we reviewed that IM manages to solve, at least partially. We will now exhibit how a developmental architecture could emerge from this work.

\begin{figure*}
\begin{center}
\includegraphics[width=15cm]{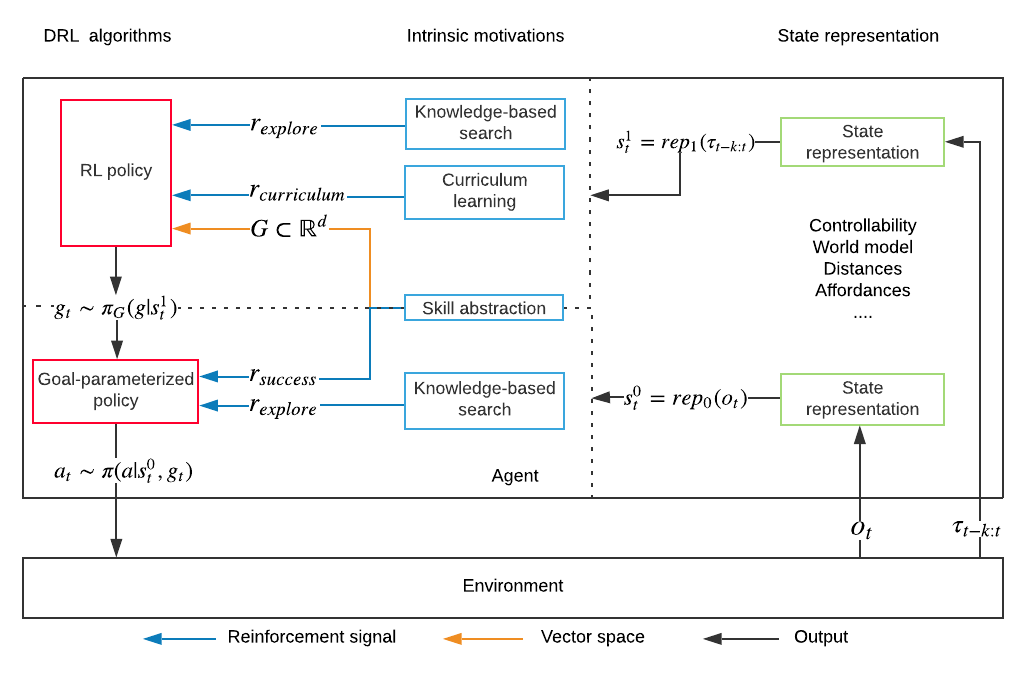}
\caption{Proposal of a developmental architecture based on several IMs, state representation and hierarchical RL. The agent observes $O_t$ at time $t$, it can compute a state representation $s_t^0$ with meaningful properties by distilling information of the world. Then it uses $s_t^0$ to compute its intrinsic rewards, which are whether its goals are achieved ($r_{success}$) or if it found something interesting ($r_{explore})$. Using these rewards, the agent can reinforce its goal-parameterized policy. If the agent terminates its option, it has to aggregate its goal trajectory $\tau_{t-k:t}$ into an upper level-specific representation $s_t^1$. Then, it can, with a standard policy algorithm, choose a goal among the goals provided by the skill abstraction module. To do so, the policy could be reinforced by knowledge-based search ($r_{explore}$) or\textit{curriculum learning}($r_{curriculum}$). We want to emphasize that we present a two-levels hierarchy of skills in our architecture for simplicity. Nevertheless, one could extend it to include as many levels as it needs by mixing curriculum reinforcement learning with skill abstraction reinforcement at intermediate levels of the hierarchy.
}

\label{im:developmental}
\end{center}
\end{figure*}

Figure \ref{im:developmental} exhibits how different works mixing DRL and IM could be integrated into a developmental architecture. We will now detail the interactions between the four intertwined components of our architecture. The first components are three intrinsic motivations which are \textit{curriculum learning}, \textit{skill abstraction}, \textit{knowledge-based search} whereas the last one is a \textit{state representation} module. The core of the potential developmental architecture could be based on \textbf{skill abstraction} (Section \ref{gen_goal}) since it encourages the agent to hierarchically build skills and represent them from scratch. It provides a goal space to the inter-option policy which can either come from the state space or have a subjective meaning, and an intra-option policy through an intrinsic reward function.
It is particularly complementary with \textbf{curriculum learning} work (section \ref{sec:curriculum}) that can accelerate the learning process and exploration but until now, mostly relies on an hand-defined goal space with only few different tasks. It results that the integration of both approaches could enable an accelerated autonomous creation of skills in an open-ended way. Ideally, both methods should be integrated in a continual learning framework \cite{parisi2019rethinking} (see also Section \ref{sec:lifelong}). We have already seen that \textit{curriculum learning} could serve to find unexplored states. However, getting new interesting skills is not obvious depending on the task environment (see \S \ref{tasks}). That is why \citename{lee2019efficient} and \citename{song2018diversity} mix exploration and skill embedding to improve the quality and diversity of skills. We think that the idea should be further explored. Typically, one could use exploration methods to find new goals, which can be to move an object or to reach an area, as illustrated by navigation tasks or manipulation tasks (see \S\ref{sec:navigation} and \S\ref{sec:manipulation}). More broadly, the agent could use any kind of \textbf{knowledge-based search}, including search for the structure of the world or controllability. 

At last, \textbf{state representation} is a critical component for all methods. In addition to speed up standard DRL algorithms, it is primordial to both explore an abstract state space and get abstract goal oriented behaviors. For example, to get object-oriented behaviors (moving toward an object for example), the agent must have notions of object in a way or an other. It can make exploration and skill acquisition a lot easier and meaningful (\S\ref{sec:staterepr}). Using a common representation across all modules at the same hierarchy level could considerably speed up the computation of both the intrinsic rewards and the reinforcement module. However a distinct representation across hierarchy levels may be necessary to reduce the complexity of computation. Typically, high-level representations may focus on position of objects whereas low-level representations should rather focus on proprioceptive data.

\citename{piaget1936naissance}, in his theory on cognitive development, argues that humans progress through four developmental stages. The first one is the sensorimotor stage which lasts from the birth of a baby to its second year. \citename{guerin2011learning} points out that the baby learns to use knowledge on the world to modify his skills. As an example, once a child understood spatial movements, he can take advantage of this knowledge to shift an object with a stick. This kind of adaptation is mainly unused in current works. Nevertheless, some exploration methods learn, for example, a large predictive model, without using this accumulated knowledge. It emphasizes the current under-exploitation of exploration methods and the lack of guidance from \textbf{knowledge} when the agent chooses skills and trains on it. More particularly, there is still no general way to distill all these discoveries in the state representation. Most information compression methods are either very specific or do not compress in a vector space. A detailed study on elements of such knowledge (intuitive physics, causality,...) can be found in \citename{lake2017building}.

To the best of our knowledge, most of previous RL-based frameworks are more specific than our developmental architecture. For example, \citename{levy2018hierarchical} propose a multi-level DRL based agent, but neither learn a \textit{state representation}, nor integrate \textit{knowledge-based} or \textit{curriculum} IMs. Closer to our proposal of architecture is the IMGEP framework, which is not restricted to RL. In this framework, an agent chooses goals with a multi-armed bandit and learns to fulfill its goal through algorithms which may or may be not from the DRL field. Interestingly, some works in the IMGEP framework already take advantage of a learned state representation to compute $r_{curriculum}$ and $r_{success}$ \cite{laversanne2018curiosity}. However we want to emphasize two major differences between the two frameworks: 1- IMGEP works alleviate the role of \textit{knowledge-based} IMs; in contrast, we emphasize the need for knowledge-based search to guide the policy at each level of a multi-level hierarchical agent, since it can help the agent to discover new goals and regularities. 2- IMGEP considers the second level of a skill hierarchy as the last goal provider. However, we believe this is important to consider a deeper skill hierarchy in order to make easier \textit{transfer learning}, \textit{exploration} and \textit{credit assignment} when an agent would have to take higher-level decision than what current tasks allow. %Typically, such very high-level decision could be \textit{going to the park}. 
A counterpart of using hierarchical reinforcement learning may be the difficulty to either choose or extend the number of levels in an automated way. This difficulty also suggests complex interactions between \textit{skill abstraction} reward $r_{success}$ and \textit{curriculum learning} reward $r_{curriculum}$ at intermediate levels (between the upper level and the lower level).

%Lastly, we believe our architecture describes the key elements of a Powerplay architecture \cite{schmidhuber2013powerplay}

\subsection{Conclusion}

To summarize, each part of the work we reviewed is related to one aspect of developmental learning. Each aspect relies on an intrinsic reward which measures the agent's ability to abstract new regularities in different parts of its trajectories. To do that, information theory is a powerful measuring tool. Theoretically linking up these different aspects brings out a developmental architecture, highlighting outlooks of the domain. In fact, concretely implementing our architecture may require to tackle the challenges we explained in Section \ref{limite}. As a consequence, such architecture may allow the agent to tackle most of the tasks we reviewed in Section \ref{tasks}. Indeed, ideally, one would like its agent to navigate through first-view vision and proprioceptive data into a very large environment, like humans do.

\section{Conclusion}

In this survey, we have presented the current challenges faced by DRL: namely 1- learning with  \textit{sparse rewards} through exploration; 2- \textit{building a hierarchy of skills} in order to make easier credit assignment, exploration with \textit{sparse rewards} and \textit{transfer learning}; 3- building a \textit{state representation} to speed up the learning process with/without access to rewards; 4- finding a \textit{curriculum} in order to improve exploration if skills are related to the state space and to allow the acquisition of several complex skills.

We identified several types of IM to tackle these issues, that we classified into two broad categories which are \textit{knowledge acquisition} and \textit{skill learning}.  

The first category refers to the agent's ability to get information on its environment, such as environment dynamics (\textit{exploration}), properties of objects (\textit{states representation}) or controllability (\textit{empowerment}). \textit{Prediction error}, \textit{novelty search} and \textit{information gain} are three competitive knowledge gain methods to improve exploration. Each one has its particularities: \textit{prediction error} methods strongly rely on a good state representation; \textit{novelty search} methods are often computationally inefficient; \textit{information gain} methods are hard to compute but elegantly avoid the \textit{white-noise} problem. In contrast to exploration, few IMs explicitly try to learn a good \textit{state representation}: some strive to find the \textit{distance between states}, and map it into the representation, others try to figure out the intrinsic \textit{disentanglement} of the world. Lastly, \textit{empowerment} characterize the search for \textit{controllability} in the state space. It can be used to substitute the extrinsic reward to get a survival mechanism or to detect elements of control.

The second category is the ability to \textit{discover} and \textit{abstract skills} in the environment. Currently, there are two broad ways to abstract skills, the first one considers every \textit{state as a potential goal}. The main issue is then to measure the similarity between states and goal-states. The second one builds skills based on a \textit{diversity heuristic}; it forces skills to lead to different parts of the state space. The major drawback is their intrinsic stochasticity and the inability to learn these skills in an end-to-end fashion. Introducing a \textit{curriculum} among the discovery of skills can considerably speed up the learning of such skills. Doing so, we found that \textit{goal sampling} methods simulate \textit{novelty search} by learning or acting on goals different with each other. \textit{Learning progress} methods allow to focus on goals at the intermediate difficulty for the agent and alleviate the need for a well-defined goal space but require a discrete goal space. \textit{Adversarial methods} can simulate \textit{learning progress} on continuous goal space but are harder to train end-to-end with a task. In fact, when skills are related to the goal space, curriculum methods may be competitive with exploration through knowledge acquisition.

Transverse to both categories, we identified and described several \textit{major challenges} and considered different ways to address them. 1- Exploration methods hardly handle stochastic environments, emphasizing the need for either a good state representation or better \textit{learning progress}/\textit{information gain} measures. 2- Knowledge-based exploration methods can be stuck when the intrinsic reward itself becomes sparse or when the environment is partially observable. If the first issue could be addressed through planning or goal-parameterized exploration, the second would certainly require to learn a state representation. 3- With knowledge-based exploration methods, the fast-moving reward function could require a better credit assignment. It suggests that using hierarchies of skill could considerably speed up exploration. 4- Building a more significant state representation could open new perspectives in standard DRL as well as in intrinsically motivated works. 5- Goal-parameterized RL currently suffers from catastrophic forgetting when there is a large number of tasks. Nevertheless, there is a large literature on models tackling catastrophic forgetting in neural networks.

Besides, a short focus on tasks addressed by works highlighted how compatible IMs and tasks are. Firstly, diversity-based skills may be adequate to get locomotion behaviours based on proprioceptive state space. Secondly, there is clearly a lack of disentangle state representation in order to learn target-oriented behaviours in both navigation and manipulation tasks. It could improve exploration through both curriculum and knowledge search. Lastly, partially observable tasks, which are the more complex, have only been tackled by few works and should be further investigated.

Ideally, one would like to learn to achieve these tasks all together. In our analysis, we identified each type of IM as a small and compatible block of a larger developmental architecture that extends previous frameworks like IMGEP. Our study also suggests that each block can be learned with a reinforcement learning algorithm and a module generating the intrinsic reward. This module tries to compress information on the basis of mutual information and general assumptions about the structure of the world. As a consequence, we show that, if associated to a module that distill knowledge into a state representation, only two principles could lead to open-ended learning agents. One could wonder what are the obstacles to the implementation of such architecture; in fact, tackling the challenges we have enumerated will be the key to concretely implement such architecture.

%\vskip 0.2in

%\bibliographystyle{splncs03.bst}
%\bibliographystyle{apalike}
\bibliographystyle{named.bst}
\bibliography{references.bib}

\end{document}